\documentclass[journal,12pt,onecolumn,draftclsnofoot,]{IEEEtran}

\IEEEoverridecommandlockouts
\usepackage{cite}
\usepackage{amsmath,amssymb,amsfonts}
\usepackage{algorithmic}
\usepackage{graphicx}
\usepackage{textcomp}
\usepackage{xcolor}
\def\BibTeX{{\rm B\kern-.05em{\sc i\kern-.025em b}\kern-.08em
    T\kern-.1667em\lower.7ex\hbox{E}\kern-.125emX}}
\usepackage{graphicx}
\usepackage{subcaption}
\graphicspath{ {images/} }
\usepackage[utf8]{inputenc}
\usepackage{fancyhdr}
\usepackage{dirtytalk}
\usepackage{tabularx}
\PassOptionsToPackage{hyphens}{url}\usepackage{hyperref}
\usepackage[ruled,lined,linesnumbered]{algorithm2e}

\usepackage{enumitem,kantlipsum}
\usepackage{bbm}
\usepackage{etoolbox} 
\patchcmd{\subsubsection}{\itshape}{\bfseries}{}{}

\usepackage{graphicx}
\graphicspath{ {images/} }
\usepackage[utf8]{inputenc}
\usepackage{fancyhdr}
\usepackage{dirtytalk}
\usepackage{tikz}
\usetikzlibrary{shapes.geometric, arrows}
\usepackage{pgfplots}
\pgfplotsset{width=8cm,compat=1.16}
\usepgfplotslibrary{external}
\usepackage{mathtools}

\newtheorem{definition}{\normalfont{\textbf{Definition}}}

\tikzstyle{startstop} = [rectangle, rounded corners, minimum width=2cm, minimum height=0.5cm,text centered, draw=black, fill=red!30]
\tikzstyle{process} = [rectangle, minimum width=2cm, minimum height=0.5cm, text centered, draw=black, fill=orange!30, align=left]
\tikzstyle{decision} = [diamond, minimum width=1.0cm, minimum height=0.4cm, text centered, draw=black, fill=green!30]
\tikzstyle{arrow} = [thick,->,>=stealth]

\newcolumntype{L}{>{$}l<{$}}

\usepackage[labelformat=simple]{subcaption}

\begin{document}

\title{Semantic Information Marketing in The Metaverse: A Learning-Based Contract Theory Framework}

\author{Ismail Lotfi, Dusit Niyato, Sumei Sun, Dong In Kim and Xuemin (Sherman) Shen

\thanks{
\textit{(Corresponding author: Dong In Kim)}

Lotfi Ismail and Dusit Niyato are with the School of Computer Science and Engineering, Nanyang Technological University, Singapore (e-mail: ismail003@e.ntu.edu.sg, dniyato@ntu.edu.sg).

Sumei Sun is with the Institute for Infocomm Research, Agency for Science, Technology, and Research (A*STAR), Singapore (e-mail: sunsm@i2r.a-star.edu.sg).

Dong In Kim is with the Department of Electrical and Computer Engineering, Sungkyunkwan University (SKKU), Suwon, South Korea(e-mail: dikim@skku.ac.kr).

Xuemin (Sherman) Shen is with the Department of Electrical and Computer
Engineering, University of Waterloo, Waterloo, ON, Canada (email: sshen@uwaterloo.ca).
}
}

\maketitle

\begin{abstract}
In this paper, we address the problem of designing incentive mechanisms by a virtual service provider (VSP) to hire sensing IoT devices to sell their sensing data to help creating and rendering the digital copy of the physical world in the Metaverse. 
Due to the limited bandwidth, we propose to use semantic extraction algorithms to reduce the delivered data by the sensing IoT devices.
Nevertheless, mechanisms to hire sensing IoT devices to share their data with the VSP and then deliver the constructed digital twin to the Metaverse users are vulnerable to adverse selection problem. The adverse selection problem, which is caused by information asymmetry between the system entities, becomes harder to solve when the private information of the different entities are multi-dimensional. We propose a novel iterative contract design and use a new variant of multi-agent reinforcement learning (MARL) to solve the modelled multi-dimensional contract problem.
To demonstrate the effectiveness of our algorithm, we conduct extensive simulations and measure several key performance metrics of the contract for the Metaverse. Our results show that our designed iterative contract is able to incentivize the participants to interact truthfully, which maximizes the profit of the VSP with minimal individual rationality (IR) and incentive compatibility (IC) violation rates. 
Furthermore, the proposed learning-based iterative contract framework has limited access to the private information of the participants, which is to the best of our knowledge, the first of its kind in addressing the problem of adverse selection in incentive mechanisms.

\end{abstract}


\section{Introduction}

Driven by the Covid-19 pandemic, the Metaverse has gained huge interest recently from different industry and public sectors~\cite{Minrui_COMST_2022, Minerva_2020}. Considered as the next generation of the Internet, the Metaverse enables users and objects to experience near real-life interaction with each other in the virtual environment through their avatars.
The Metaverse is made up from different emerging technologies such as virtual reality (VR), augmented reality (AR) and haptic sensors.
Furthermore, other emerging technologies such as beyond 5G and 6G are driving the Metaverse from imagination and fiction towards real world implementation as they enable users to access the Metaverse from anywhere, anytime instantly.

The first step towards realizing and exploiting the Metaverse is the replication of the physical objects into their respective digital twins. As the digital twins are required to replicate the physical real-world system to the finest details~\cite{Minerva_2020}, generating an accurate 3D model of the physical system and constant update of the physical system in digital twin is the first step towards this goal.
However, the creation of an accurate 3D digital copy is challenging for several reasons. 
First, in the upstream layer, i.e., between the VSP and the sensing IoT devices, the collected data by the sensing IoT devices is huge in size and the available bandwidth for data transmission will quickly exceed the system limitation. In addition, the delivered data by the sensing IoT devices needs to be delivered timely and should not be outdated.
Second, in the downstream layer, i.e., between the VSP and the Metaverse users, to support a real-time interaction between the Metaverse users and the physical world, the rendered digital twin by the VSP needs to be delivered timely and with an acceptable quality to the Metaverse users.
The quality of the digital twin can be defined based on the resolution and frequency of the received video fragments~\cite{Minrui_COMST_2022}.
Therefore, to enable a real-time construction and delivery of the digital twin in the Metaverse, the communication system needs to be carefully designed so as to maximize successful data transmission with high data value while minimizing the latency of packet delivery.

However, it is challenging for the VSP to find a balance between the revenue, i.e., profit, and the cost in both layers. In the upstream layer, the VSP needs to find an optimal strategy to maximize its revenue while minimizing the cost, i.e., prices, given to the sensing IoT devices for their data. In the downstream layer, the VSP also needs to find another optimal strategy for the prices to sell the digital twin service to the Metaverse users with the constraint on the delivery costs, i.e., computation cost for rendering and delivering the digital twin to the Metaverse users.
The VSP needs to design incentive mechanism in both layers to motivate the sensing IoT devices and the users to participate in the Metaverse ecosystem. Specifically, the VSP uses private information of the participants to design a mechanism that satisfies the incentive compatibility (IC) and individual rationality (IR) properties for each participant.
However, the uncertainty of the VSP about private information of the participants further hinders the derivation of the optimal strategy, causing the problem of information asymmetry~\cite{Bolton_2005}.
The problem of information asymmetry encourages malicious participants to misreport their private information truthfully to gain higher profits.
For instance, some sensing IoT devices with low ability to provide rich semantic information and fresh data might claim to have a higher level than their true type, causing the VSP to provide them with payments higher than what they truly deserve.
This behaviour can similarly happen when the VSP intends to deliver the digital twin to the Metaverse users. The Metaverse users can misreport their private valuation about the delivered digital twin (which are based on their private types) to push the VSP to decrease the offered prices and hence, getting a higher utility than deserved. 

To address the aforementioned challenges, we first propose the use of semantic information extraction algorithm on the raw data collected by the sensors on the IoT devices to minimize the size of the transmitted data~\cite{Ouaknine_2021_ICCV}. Instead of transmitting the raw data to the VSP, the IoT devices transfer only relevant information (semantic) which can be used directly to create a 3D copy of the physical world. To formulate our problem, we then adopt contract theory which is regarded as an efficient tool to design incentive mechanism under asymmetric information scenarios~\cite{Bolton_2005}.
Nevertheless, to properly design the contract and maximize its revenue, the VSP needs to be able to categorize the users based on their private types which can be multi-dimensional.
In the upstream layer, different sensing IoT devices can have different semantic information abilities, different speed for data delivery. Similarly, in the downstream layer, different Metaverse users can have different private valuation towards different qualities of the digital twin delivered by the VSP, e.g., resolution and refresh rate. Therefore, the VSP needs to take these multi-dimensional private information of the participants of the Metaverse ecosystem while designing the incentive mechanism.

However, solving multi-dimensional contracts is not straightforward to address.
Existing works proposed to reduce the multi-dimensional contract into a single dimensional contract~\cite{Zhiyuan_2019_TMC, Xiong_2020_TWC_Contract}. The derived single dimensional contract is then solved by using standard single-crossing condition technique. However, this conversion requires tedious mathematical transformations with proofs for IC and IR properties in addition to the computation efficiency. For instance, the single-crossing condition does not always hold when there are more than one type~\cite{Zhiyuan_2019_TMC, Xiong_2020_TWC_Contract}. Moreover, if the definition of the utility function of either the contract designer or the users changes slightly, the derived solution needs to be reformulated from scratch. Furthermore, all of the existing techniques to solve the single-dimensional contract requires certain assumptions about the used functions, e.g., monotonicity, which adds more limitations for the generality of the derived solutions. 
In this work, we address the problem of solving multi-dimensional contract from a totally different perspective. This is a key contribution of this paper. Specifically, we formulate the contract problem as a Markov Decision Process (MDP) and solve it using a new variant of multi-agent reinforcement learning (MARL) algorithm.

To the best of our knowledge, no previous work has attempted to solve the contract optimization problem using DRL. This is due to the fact that the nature of the problems addressed by DRL is different from those of contracts. For instance, a closely related work was proposed in~\cite{Minrui_ICC_2022} in which the authors adopted the DRL framework to solve the double Dutch auction (DDA). The DDA is conducted in many rounds and the DRL is suitable to learn the optimal ``step size" over time.
However, in contract, the contract bundles are delivered only once to the participants, and the participants select their preferred bundles only once. A major challenge for using DRL in contract is the generation of the training set. Typically, in problems where DRL is applied, the environment is highly repetitive, i.e., decisions are taken frequently, which enables the collection and evaluation of previous actions. However, in contract, the contract bundles are generated only once and they need to guarantee the optimality in addition to the IC and IR properties for all participants. 
These system settings make it non-trivial to adopt DRL for our problem.
Nevertheless, motivated by the DRL's generality, we are able to create a learning environment for the proposed iterative contract problem. 

In this paper, we extend our previous work in~\cite{Ismail_2022_FNWF} to a multi-dimensional asymmetric information problem in a two-layer Metaverse system and develop a learning-based iterative contract to solve it. In summary, the main contributions of our work are as follows:

\begin{itemize}
    \item We design a novel two layer Metaverse ecosystem where in the first layer, the VSP hires sensing IoT devices to collect data from the physical world, while in the second layer the VSP uses the collected data to create the digital twin of the physical world and delivers it to the Metaverse users. To minimize the data volume over the wireless link, we require the sensing IoT devices to extract and transmit only the semantic information from the raw data. The proposed design is shown to achieve the objectives of the Metaverse ecosystem, i.e., fast delivery and update of reliable information.

    \item We then use the contract theory framework to design an incentive mechanism to incentivize the participants in both layers, i.e., sensing IoT devices and Metaverse users, to engage in the Metaverse ecosystem and mitigate the adverse selection problem. We propose a novel iterative contract framework to solve the challenging multi-dimensional optimization problem. To the best of our knowledge, this is the first work that applies contract theory in a two-layer Metaverse system. It is non-trivial to design an incentive mechanism for such systems due to information asymmetry at different layers, i.e., the data collection layer and data delivery layer.

    \item To solve the resultant iterative contract model, we develop a new variant of MARL systems where we consider that the VSP creates instances for each participant in the contract and interact with each other until reaching a feasible solution that maximizes the profit of the VSP while minimizing the IR and IC violation rates. In other words, the VSP is the only entity that keeps changing the prices for its bundles and observes how the participants choose their optimal bundles. The VSP also receives information about how many IR and IC violations have occurred in each round. This information is considered to be available for all the participants during the learning process by augmenting the MDP's observation space to cover all that of the other participants. To the best of our knowledge, this MARL design is the first of its kind.
\end{itemize}

The structure of the paper is as follows. In Section~\ref{section_SystemModel} we define our system model and provide some preliminaries about semantic information for the Metaverse. In Section~\ref{section_contractFormulation} we formulated the optimization problem as a contract theory problem and develop our learning-based iterative contract model. Finally, we provide numerical results and insightful discussions about our framework in Section~\ref{section_Results}. Section~\ref{section_conclusion} concludes the paper.

\section{Related Works}

\subsection{Metaverse Services}

Digital twin modeling of the physical world in the Metaverse has a number of benefits for different application scenarios.
In~\cite{Minrui_COMST_2022}, the authors provided a detailed survey about the Metaverse and its applications and challenges from a communication perspectives.
As the study of the Metaverse is still in its infancy, only few works have addressed the aspect of wireless resource allocation for the Metaverse. In~\cite{Han_ICC_2022}, an IoT-assisted Metaverse sync problem was studied in which an evolutionary game was formulated to enable the IoT devices to select VSPs to work for. However, the volume of the data and the limited bandwidth problem was not addressed. In~\cite{Hashash_2022}, an iterative algorithm based on transport theory was used to minimize the delivery time between the sensing IoT devices and the VSP, and hence minimize the gap between the real-time state of the physical twin and the state of the digital twin.  However, it was assumed that the computation time, bandwidth, rate of the sensing IoT devices are precisely known. However, the sensing IoT devices might belong to different entities and hence, might misreport their private information and cause the whole Metaverse system to collapse. Therefore, there is a need to design incentive mechanism to ``incentivize" participating sensing IoT devices to report their private information truthfully.

\subsection{Semantic Communication for The Metaverse}
Semantic communication for the Metaverse can be enabled at two different layers. The first layer is the physical layer where the objective is to use some technique, e.g., machine learning (ML), to reduce the number of transmitted bits between the transmitter and the receiver. The second layer is at the application layer where the raw data is reduced in size through ML or other techniques by extracting only the valuable, i.e., semantic, information. The derived semantic information is then transmitted to the receiver through standard wireless transmission technique.
In~\cite{Park_arxiv_2022}, the authors developed a semantic multiverse communication system using generative adversarial networks (GANs). The encoder learns the semantic representations of the data, while the generator learns how to manipulate the extracted semantics for locally rendering the digital twin in the Metaverse.
In~\cite{Jiacheng_TWC_2022}, authors used contest theory to design a semantic-aware sensing information transmission for the Metaverse.
Causality was used in~\cite{ThomasChristo_ARXIV_2022} to infer cause-effects relationship between the transmitted symbols and the received ones over the wireless communication channel. A game theory based language model was then developed to enable minimalist representation and transmission of the extracted semantics.
In~\cite{tang2019_learning}, the relationship between objects in images was represented as a graph, and in~\cite{Krishna_2016_VGenome} a dataset for this purpose was presented. In our recent work~\cite{Ismail_2022_FNWF}, we used the idea presented in~\cite{tang2019_learning} and proposed an application-layer semantic communication system for the Metaverse. A reverse auction mechanism was then developed to incentivize the data owners, i.e., sensing IoT devices, to make their bids to sell semantic information to the VSP truthfully.

\subsection{Multi-Dimensional Contracts}
To derive the optimal pricing bundles, the system designer needs to have full knowledge about the private information of the participants. To incentivize the participants to reveal their private information, existing works used either Stackelberg games or contract theory. However, the shortcoming of Stackelberg games is that they can be used only for scenarios with a single-dimension private type~\cite{liu_2022contract}. Therefore, several works used the framework of contract theory to design incentive mechanisms for problems with multi-dimensional private types. In the area of wireless communications, the idea of extending single-dimensional contract to two-dimensional contract was first proposed in~\cite{Zhiyuan_2019_TMC}, where the authors proposed to use an additional auxiliary variable to transform the two dimensional contract into a single-dimensional contract. The derived contract was then solved by using standard approaches in contract theory to prove the IC and IR properties in addition to the optimality of the derived solution.
Motivated by~\cite{Zhiyuan_2019_TMC}, the authors in~\cite{Xiong_2020_TWC_Contract} extended the idea to a three-dimensional contract. 
However, this approach requires tedious formulation of the problem and several assumptions about the system dynamics to prove that the designed contract is truthful, i.e., does not violate the IC and IR properties.
Finally, some existing works used the framework of contract theory to improve the performance of some ML problems, e.g., federated learning as in~\cite{Jiawen_JIOT_2019_Contract_FL}. However, to the best of our knowledge, no prior work has attempted to use an ML technique, e.g., DRL, to address the aforementioned challenges in contract designs.

Motivated by the limitations mentioned above, we propose a learning-based iterative contract to derive the optimal pricing for the participants in the Metaverse ecosystem.
Therefore, our studied Metaverse system can be regarded as an instance of a general framework for solving multi-dimensional contracts using DRL.

\section{System Model And Preliminaries}\label{section_SystemModel}
We consider a digital market consisting of data owners, a VSP and Metaverse users. The data owners, e.g., IoT devices equipped with sensors, collect data about the physical environment and sell it to the VSP. The VSP then creates the digital twin of the physical environment and commercializes the digital twin to different Metaverse users. A two-layer contract theory-based framework is developed for the VSP to determine prices for purchasing data from the sensing IoT devices and for selling digital twin to the Metaverse users.

\subsection{Metaverse Platform}

\begin{figure}[ht!]
    \centering
    \includegraphics[width=.45\textwidth,height=4.5cm]{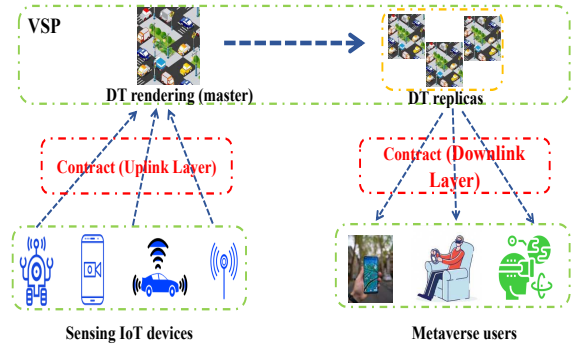}
    \caption{System model.}
    \label{fig:sys_model}
\end{figure}

As illustrated in Fig.~\ref{fig:sys_model}, we consider a VSP that is collecting sensing data from a set of IoT devices, denoted as $\mathcal{N} = \{1, \dots, N\}$ in the filed, e.g., vehicles or smartphones. The edge server, which is monitored by the VSP, is responsible for the replication of the physical twin by rendering the received data into a digital twin (DT) of the physical twin.
Next, the VSP sends the digital twin to the set of Metaverse users, defined as $\mathcal{M} = \{1, \dots, M\}$.
Each sensing IoT device has a set of sensors to collect geo-spatial data from the surrounding environment and send the data back to the VSP. However, raw data is usually large in size adding further limitations on the required bandwidth and data delivery latency. Therefore, the IoT devices are equipped with machine learning (ML) models to extract only the semantic information from the collected raw data, which is smaller in size, and send the semantic information to the VSP.
Nonetheless, if the received data from the IoT devices is outdated, the created digital twin will not be able to reflect real time dynamics of the physical twin. Therefore, the VSP leverages an age of information (AoI) metric to measure and guarantee freshness of the received data from the IoT devices.
Once all of the semantic information is received by the VSP, the digital twin is created and distributed to the Metaverse users.
In what follows, we discuss preliminaries about semantic information, AoI and their roles in deriving the value of the collected information by the sensing IoT devices and then we describe the delivery model of the digital twin by the VSP to the Metaverse users.

\subsection{Fresh Semantic Information Collection Model}
\subsubsection{Sensing IoT devices Modeling}
Different from traditional crowd-sensing platforms that collect all raw data from data owners directly, the VSP obtains only semantic information from the IoT devices (e.g., semantic mask for each object in an image with its corresponding class or semantic text from voice recording). 
The incorporation of semantic information into our system is motivated by the following reasons:
\begin{itemize}
    \item The number of communication channels available to the VSP are limited. Hence, if the VSP allows transmission of raw data by the IoT devices, only few devices will be able to transmit their data which reduces the heterogeneity of the collected data.
    
    \item Raw data is large in size in general (e.g., video and images), which can increase the transmission delay, making the rendering of the digital twin very slow and obsolete. 

    \item The quality of the constructed digital twin will be higher as more semantic information about the physical world will be available to the VSP.
\end{itemize}

Let $\Psi = \{\psi_e : e \in \{1,\ldots,E\}\}$ denote the set of different semantic levels (or scores) available.
The similarity score is impacted mainly by the algorithm used by the sensing IoT devices for semantic information extraction as demonstrated in~\cite{Ismail_2022_FNWF, Ouaknine_2021_ICCV}.
We consider the algorithms as types (integers) and they are sorted in an ascending order, i.e., $0 < \psi_1 \leq \psi_2 \leq \dots \leq \psi_E$. Note that the semantic extraction algorithm used by each sensing IoT device depends on the types of sensors equipped in each IoT device, e.g., camera and radar.
Typically, sensing IoT devices with a high semantic score value can provide the VSP with more accurate and rich set of information. Therefore, they are more preferred by the VSP and should receive more payment for their data. 
However, as the sensing IoT devices are owned by independent parties, their capabilities of extracting the semantic information is different, heterogeneous and private. For instance, two IoT devices might have the same price for selling their semantic information, and the VSP might be indifferent when choosing which IoT device to buy data from. Therefore, if the VSP is aware of the semantic value of each IoT device, i.e., the ability of the IoT device to extract more accurate semantic information, the VSP can then choose the IoT devices that increase the quality of its constructed digital twin. 

Nevertheless, the provided semantic information is affected directly by the reliability of the network link between the sensing IoT devices and the VSP. Even if the value of the provided semantic information is high, the link with high bit error rate (BER) can prevent the VSP from receiving the extracted semantic information about the physical world efficiently, and hence making the rendering at the Metaverse obsolete.
To mitigate this issue, we consider the radio link transmission rate as a valuation metric for the link quality. The transmission rate can be adjusted by the transmitters through allocating more channels and/or increasing the transmit power to increase the signal-to-noise ratio (SNR) at the receiver.
In what follows, we denote $\Lambda = \{ \lambda_b: b \in \{1,\ldots,B\}\}$ to be the set of different available transmission rate values which are also sorted in an ascending order\footnote{Note here that the transmission rate has a discrete value due to modulation and coding schemes.}, i.e., $0 < \lambda_1 \leq \lambda_2 \leq \dots \leq \lambda_B$.

We also consider the age of information (AoI), which is defined as the time elapsed since the generation of the last received data at the source, as an important criterion of the delivered semantic information.
Specifically, due to the congestion at the transmission queues, the AoI is directly impacted and becomes larger. 
It has been shown in~\cite{Sanjit_AoI_2012} that with \emph{first-come-first-served} (FCFS) queues, increasing the refresh rate does not yield a small AoI as this strategy may lead the destination to receive delayed status update because the packets become backlogged in the communication system. 
These reasons motivates the use of \emph{last-come-first-served} (LCFS) queues with preemption as in~\cite{Sanjit_AoI_2012}. Under LCFS with preemption, the new generated packet is allowed to replace the current packet in service and hence, maintain a low AoI.
Let $\Gamma = \{\gamma_c : c \in \{1,\ldots,C\}\}$ denote the set of different refresh rate types. The refresh rate type is related to the average AoI $\varpi_{\gamma}$ at the VSP as follows~\cite{Sanjit_AoI_2012}
\begin{equation}\label{AoI_varpi}
    \varpi_{\gamma} = \frac{1}{\gamma} + \frac{1}{\mu},
\end{equation}

\noindent where $1/\gamma$ is the mean packet arrival time at the VSP and $1/\mu$ is the mean processing time at the VSP server. From~\eqref{AoI_varpi} we observe that when $\mu$ is considered constant the average AoI is inversely proportional to the refresh rate $\gamma$. Therefore, sensing IoT devices which have higher refresh rates bring more utility to the VSP. 
Based on the findings in~\cite{Yates_2018_ISIT, Bedewy_2019}, we consider that as the refresh rate increases, the AoI decreases following a non-increasing convex function.
Without loss of generality we consider that refresh rate types are sorted in an ascending order similar to the other types, i.e., $0 < \gamma_1 \leq \gamma_2 \leq \dots \leq \gamma_C$. 

In short, each sensing IoT device is differentiated by its three dimensional private information: the semantic score value $\psi_e$, transmission rate value $\lambda_b$ and refresh rate value $\gamma_c$\footnote{Note here that the transmission rate captures the volume of data that can be transmitted over a period of time while the refresh rate captures the number of times the transmitter sends a new update to the receiver.}.
Therefore, the utility of a sensing IoT device with type-($\lambda,\gamma,\psi$) to deliver semantic information to the VSP with volume size $\hat{s}$ is defined as
\begin{equation}
    U^{\dagger}(\hat{s}_{\lambda,\gamma,\psi}) = \pi^{\dagger}_{\lambda,\gamma,\psi} - \Upsilon^{\dagger}(\hat{s}_{\lambda,\gamma,\psi}),
\end{equation}

\noindent where $\pi^{\dagger}_{\lambda,\gamma,\psi}$ is the price associated to data with quality $\hat{s}_{\lambda,\gamma,\psi}$,
$\Upsilon^{\dagger}(\hat{s}_{\lambda,\gamma,\psi})$ refers to the cost for delivering a data with size $\hat{s}$ by an IoT device with type-($\lambda,\gamma,\psi$) to the VSP, and is defined as
\begin{equation}
    \Upsilon^{\dagger}(\hat{s}_{\lambda,\gamma,\psi}) = \Upsilon^{\dagger}_0 + T^{\dagger}(\hat{s}_{\lambda,\gamma,\psi}), 
\end{equation}

\noindent where $\Upsilon^{\dagger}_0$ is a fixed cost and $T^{\dagger}(\hat{s}_{\lambda,\gamma,\psi})$ is the specific cost for data generated by type-($\lambda,\gamma,\psi$) IoT device. The cost function reflects both the computation cost (i.e., data collection and semantic information extraction) and communication cost (i.e., channel allocation by the sensing IoT devices).

\subsubsection{VSP modeling}
The VSP needs to properly design rewards for each sensing IoT device type. 
Different from most existing works where some private information of the users are known to the VSP, we are considering a more realistic asymmetric information scenario in which all the private information of the sensing IoT devices are not known to the VSP. 
In other words, the VSP does not know exactly the type of each IoT device, i.e., its semantic score value $\psi_e$, its transmission rate value $\lambda_b$ or its refresh rate value $\gamma_c$. 
To solve the asymmetric information problem, we incorporate contract theory into our model. Specifically, the VSP designs specific contract bundles for each type of the sensing IoT devices with the aim of maximizing its utility, i.e., profit, while ensuring that each sensing IoT device does not deviate from choosing the bundle designed for its true type.
The VSP designs a contract bundle, denoted as $\Omega^{\dagger} = \{\omega_{\lambda,\gamma,\psi}: \lambda \in \Lambda,\gamma \in \Gamma,\psi \in \Psi\}$ that consists of $B\times C\times E$ contract items denoted as $\omega_{\lambda,\gamma,\psi}=\{\hat{s}_{\lambda,\gamma,\psi}, \pi^{\dagger}_{\lambda,\gamma,\psi} \}$ and characterized by a joint probability mass function $Q^{\dagger}(\lambda,\gamma,\psi)$ for each IoT device's joint type combination. Hence, the VSP's utility from type-($\lambda,\gamma,\psi$) IoT device is given by
\begin{equation}\label{eq:R_ddagger}
    R^{\dagger}(\omega_{\lambda,\gamma,\psi}) = \sigma f(\hat{s}_{\lambda,\gamma,\psi}) - \pi^{\dagger}_{\lambda,\gamma,\psi} +(K - \varpi_{\gamma}),
\end{equation}

\noindent where $\sigma$ is the revenue coefficient for the VSP and $\sigma f(\hat{s})$ is the revenue of the VSP from the data received from the sensing IoT device with type-(${\lambda,\gamma,\psi}$). 
Motivated by~\cite{Zehui_2020_TMC}, we adopt the $\alpha$-fairness function to define $f(\hat{s})$ as follows:
\begin{equation}\label{eq:alpha_fairness_func}
    f(\hat{s}) = \frac{1}{1-\alpha} \hat{s}^{1-\alpha} ,
\end{equation}

\noindent where $0 <\alpha<1$ is a given constant.
The last term $(K - \varpi_{\gamma})$ in~\eqref{eq:R_ddagger} represents the benefit from the AoI. Specifically, $K$ is a constant and $(K - \varpi_{\gamma})$ can be interpreted as a satisfactory function from the average AoI~\cite{Zhou_Xuying_TCOMM_2021}, where a low average AoI brings a high benefit to the VSP.
The overall utility of the VSP from all sensing IoT devices is then formulated as
\begin{equation}
\begin{split}\label{eq:R_dagger_final}
        R^{\dagger}(\Omega^{\dagger}) = \sum\limits_{\lambda\in\Lambda}\sum\limits_{\gamma\in\Gamma}\sum\limits_{\psi\in\Psi} N Q^{\dagger}(\lambda,\gamma,\psi) \Bigl(
        \sigma f(\hat{s}_{\lambda,\gamma,\psi}) 
        - \pi^{\dagger}_{\lambda,\gamma,\psi}
        +(K - \varpi_{\gamma})\Bigr).
\end{split}
\end{equation}

\subsection{Digital Twin Delivery Model}

\subsubsection{Metaverse Users Modeling}
In our model, we define the quality of a digital twin with respect to the Metaverse users in terms of the resolution of the digital twin and the refresh rate per time unit~\cite{Bado_2022}.
The resolution captures the size of the transmitted data while the refresh rate captures the freshness of the data.
In other words, if a Metaverse user subscribes to a digital twin delivery service with resolution $r$ (e.g., pixel per inch), and refresh rate $h$ (e.g., frame per second (FPS)), the VSP will assert the delivery of the digital twin as requested to the Metaverse user. If the Metaverse user accepts to buy a replica of the digital twin with quality $(r,h)$, the VSP delivers that replica to the Metaverse user and charges with price $\pi^{\ddagger}(r,h)$. 
Nonetheless, the Metaverse users have different preferences towards various combinations of resolutions and refresh rates. To present this preference, we use a valuation function with both resolution and refresh rate parameters.
Specifically, each Metaverse user has some private valuation of both resolution and refresh rate, denoted as $\tau$ and $\phi$, respectively.
These private valuation parameters capture both \emph{resolution sensitivity}, i.e., perception, and \emph{refresh rate sensitivity}, i.e., timeliness.
Based on the works in~\cite{SIGMETRICS_2015, Zehui_2020_TMC}, we define the valuation of the Metaverse user with type-($\tau,\phi$) to the provided digital twin with resolution $r$ and refresh rate $h$ as
\begin{equation}\label{eq:V_q_tau_phi}
    V^{\ddagger}(\tau, \phi, r, h) = \tau g_1(r) + \phi g_2(h),
\end{equation}

\noindent where $g_1(\cdot)$ and $g_2(\cdot)$ follow an $\alpha$-fairness function as earlier described in~\eqref{eq:alpha_fairness_func} with changes only to parameter $\alpha$. 
The Metaverse user is also required to have enough bandwidth to receive data from the VSP in addition to its internal hardware specifications, e.g., screen refresh rate~\cite{SunYi_2016}. 
Moreover, as the refresh rate $h$ (in FPS) increases, the inter-frame time decreases which is more preferred by the Metaverse user.
The Metaverse user needs to trade-off between the quality of the delivered digital twin and the cost.
Therefore, the utility of the Metaverse user with type-($\tau, \phi$) after purchasing a digital twin with quality $(r,h)$ is defined as
\begin{equation}\label{eq:Utility_MV}
    U^{\ddagger}(\tau, \phi, r, h) =  V^{\ddagger}(\tau, \phi, r, h)  - \pi^{\ddagger}(r,h).
\end{equation}

\subsubsection{VSP Modeling}

To guarantee the delivery of the digital twin with the specified quality, the VSP needs to use a certain number of resources and algorithms which increases the delivery cost as the quality increases. 
For example, instead of using a single processor or a single queue, to deliver all the digital twin packets to the Metaverse users, the waiting time in the queue can be minimized for each Metaverse user, and hence, minimizing the AoI at the Metaverse user side~\cite{Bedewy_2019}. This adjustment significantly minimizes the AoI at the Metaverse user side but increases the cost for the VSP. 
We define the cost to the VSP to deliver the digital twin with quality $(r,h)$ as
\begin{equation}
    \Upsilon^{\ddagger}(r,h) = \Upsilon^{\ddagger}_0 + T^{\ddagger}(r,h), 
\end{equation}

\noindent where $\Upsilon^{\ddagger}_0$ is a fixed cost for the VSP to collect data from the sensing IoT devices and render the digital twin. $T^{\ddagger}(r,h)$ is the specific cost for quality $(r,h)$.
Finally, the utility of the VSP for delivering a digital twin with quality $(r,h)$ is defined as the difference between the selling price and the cost, i.e.,
\begin{equation}\label{eq:fedddfdsddd}
    R^{\ddagger}(r,h) =  \pi^{\ddagger}(r,h) - \Upsilon^{\ddagger}(r,h).
\end{equation}

\section{Contract Formulation}\label{section_contractFormulation}
In this section, we formulate the contract design problem to maximize the utility of the VSP when buying the semantic information from the sensing IoT devices and when selling the constructed digital twin to the Metaverse users.
For the contract to be feasible, it has to guarantee both the incentive compatibility (IC) and individual rationality (IR) properties for all types~\cite{Gao_2011}. In what follows, we describe IR and IC properties with respect to the upstream layer, i.e., for the contract between the VSP and the sensing IoT devices, and with respect to the downstream layer, i.e., between the VSP and the Metaverse users. Finally, we propose a DRL-based model to solve the contracts of the upstream and downstream layers, which we call \emph{iterative contract} and is -to the best of our knowledge- an unprecedented method to solve contracts.

\subsection{Upstream Layer (VSP and Sensing IoT devices)}
The VSP obtains historical data about the semantic levels and transmission rates of different sensing IoT devices.
The average AoI for each IoT device (and hence, the refresh rate) is derived by the VSP from historical interactions.
The VSP then designs a contract by solving problem~\eqref{eq:optz_UppLinkLayer} and broadcasts the designed contract to the IoT devices. Next, each IoT device sends its selected contract item to the VSP, i.e., signs the contract with the VSP. Finally, the IoT devices send their semantic information to the VSP and receive payments as specified in the contract. 
A feasible contract in an open market must satisfy the IR and IC properties.
The IR and IC properties of the upstream layer are defined as follows.

\begin{definition}
\textit{Individual Rationality (IR) for IoT device: An IoT device with type-($\lambda,\gamma,\psi$) will only accept to sell its semantic information to the VSP if its utility is non-negative, i.e.,}
\begin{equation}\label{eq:IR_IoT}
\begin{split}
    U^{\dagger}_{\lambda,\gamma,\psi}(\hat{s}_{\lambda,\gamma,\psi}) \geq 0, \quad \forall \lambda \in \Lambda, \forall \gamma \in \Gamma,\forall \psi \in \Psi.
\end{split}
\end{equation}
\end{definition}

\begin{definition}
\textit{Incentive Compatibility (IC) for IoT device:
The utility of an IoT device with type-($\lambda,\gamma,\psi$) is maximized only when selecting the contract designed for its true type, i.e.,
}
\begin{equation}\label{eq:IC_IoT}
\begin{split}
    U^{\dagger}_{\lambda,\gamma,\psi}(\hat{s}_{\lambda,\gamma,\psi}) \geq
    U^{\dagger}_{\lambda,\gamma,\psi}(\hat{s}_{\lambda',\gamma',\psi'}), \quad
    \quad \forall \lambda,\lambda' \in \Lambda, \\ 
    \forall \gamma,\gamma' \in \Gamma,\forall \psi,\psi' \in \Psi, \lambda\neq\lambda', \gamma\neq\gamma', \psi\neq\psi'.
\end{split}
\end{equation}
\end{definition}

The IR condition ensures the participation of the sensing IoT devices while the IC condition ensures that each sensing IoT device selects the contract designed for its true type. The aim of the VSP is to design a contract $\boldsymbol{(\hat{s}, \pi^{\dagger})}$ to maximize its utility taking into account the IR and IC conditions, which is expressed as follows:

\begin{subequations}
\label{eq:optz_UppLinkLayer}
\begin{align}
\begin{split}
\mathcal{P}_1:\quad \max_{\boldsymbol{(\hat{s}, \pi^{\dagger})}} 
    \sum\limits_{\lambda\in\Lambda}\sum\limits_{\gamma\in\Gamma}\sum\limits_{\psi\in\Psi} N Q^{\dagger}(\lambda,\gamma,\psi) \Bigl(
        \sigma f(\hat{s}_{\lambda,\gamma,\psi}) 
        - \pi^{\dagger}_{\lambda,\gamma,\psi}
        +(K - \varpi_{\gamma})\Bigr)
\end{split}\\
\begin{split}
\hspace{4cm} s.t. \quad  \eqref{eq:IR_IoT}~and~ \eqref{eq:IC_IoT}. \label{eq:MaxB}
\end{split}
\end{align}
\end{subequations}

However, the parameters in~\eqref{eq:IR_IoT} and~\eqref{eq:IC_IoT} are private to the sensing IoT devices and can be misreported to gain higher utility than deserved. 
Moreover, to solve $\mathcal{P}_1$ we need to address $B\times C\times E$ IR constraints and $(B\times C\times E)\times(B\times C\times E -1)$ IC constraints, which are all non-convex. Intuitively, such an optimization problem is not straightforward to solve.
The classical approach is to first define some lemmas to constrain the pricing function and the types, e.g., monotonicity and pairwise incentive compatibility, and then relax the optimization problem to reduce its complexity.
However, these methods are not directly applicable here due to the multi-dimensionality of the contract. 
Interestingly, some recent works proposed to introduce an auxiliary type to reduce the dimensionality of the contract and then solve the relaxed problem using dynamic programming or branch and bound techniques~\cite{Zhiyuan_2019_TMC, Xiong_2020_TWC_Contract}. However, these approaches add another layer of difficulty to the problem formulation, which become more tedious, time consuming to adjust and to prove the corresponding lemmas and theorems.
Furthermore, the necessary and sufficient conditions for these approaches further tighten the overall assumptions in the system and limit its generality.
In this work, we design a DRL-based iterative multi-dimensional contract that is executed over several interactions between the VSP and the sensing IoT devices. The VSP starts with a set of random bundles and converges to the optimal set of bundles, which is the objective of the contract designer. 
In what follows, we first start by defining the MDP of the upstream layer, then briefly discuss how this MDP is solved.

\textbf{Markov Decision Process:}
An MDP is defined by a tuple $<\mathcal{S}^{\dagger}, \mathcal{A}^{\dagger}, \mathcal{P}^{\dagger}, r^{\dagger}>$ where $\mathcal{S}^{\dagger}$ is the state space, $\mathcal{A}^{\dagger}$ is the action space, $\mathcal{P}^{\dagger}$ is the state transition probabilities and $r^{\dagger}$ is the immediate reward received by the agent, i.e., the VSP, after performing action $a^{\dagger}$ at state $s^{\dagger}$.

\textit{State Space:}
The state space of the system at time slot $t~(t=1,2,\dots,T)$ is defined as
\begin{equation}\label{eq:state_space_UplinkLayer}
\begin{split}
\mathcal {S}^{\dagger(t)} \triangleq \Big \{ \mathbf{a}^{\dagger(t-1)}, \boldsymbol{\pi}^{\dagger(t)}, \boldsymbol{\hat{s}}^{(t)},
\mathbf{x}^{\dagger(t)}, \mathbf{y}^{\dagger(t)} \Big \},
\end{split}
\end{equation}

\noindent where $\mathbf{a}^{\dagger(t-1)}$ is the action vector from the previous time slot, $\boldsymbol{\pi}^{\dagger(t)}$ is the price vector at time slot $t$ and $\boldsymbol{\hat{s}}^{(t)}$ is the semantic information size vector at time slot $t$. $\mathbf{x}^{\dagger(t)}$ and $\mathbf{y}^{\dagger(t)}$ are binary vectors of sensing IoT devices which have their IR an IC violated, respectively.
The system state is then defined as a composite variable $\mathbf{s}^{\dagger} \triangleq (\mathbf{a}^{\dagger}, \boldsymbol{\pi}^{\dagger}, \boldsymbol{\hat{s}},
\mathbf{x}^{\dagger}, \mathbf{y}^{\dagger}) \in \mathcal{S}^{\dagger}$.

\textit{Action Space:}
For better budget allocation, the VSP is able to dynamically adjust the prices and the semantic information size values for each contract bundle. Let $price_k$ denote the price for bundle $k$ and $\eta_{1,k}$ a scalar to adjust the price between two time slots for the contract bundle $k$. The price is updated as


\begin{equation}
    \pi^{\dagger,(t+1)}_k = price_k \times (1+\eta^{(t)}_{1,k}),
\end{equation}

\noindent where $\eta^{(t)}_{1,k} \in [-range, range]$ and $0\leq range \leq 1$. The semantic information size values are adjusted similarly.  
Let $size_k$ denote the semantic information size value for bundle $k$ and $\eta_{2,k}$ a scalar to adjust the size between two time slots for the contract bundle $k$. The semantic information size value is updated as
\begin{equation}
    \hat{s}^{(t+1)}_k = size_k \times (1+\eta^{(t)}_{2,k}),
\end{equation}

\noindent where $\eta^{(t)}_{2,k} \in [-range, range]$. 
Based on these definitions of the price and semantic information size adjustments, the action space of the VSP consists of the joint action of reducing, increasing or keeping the current price and semantic information size value for all contract bundles at time slot $t$.
Therefore, the action space is defined by: $\mathcal{A}^{\dagger} \triangleq \Big \{ (a', a'') : a', a'' \in~\{0, 1, 2\}  \Big \}$,
\noindent where $a'=0$, $a'=1$ and $a'=2$ refer to the actions of increasing the semantic information size, decreasing the semantic information size or keeping the current size, respectively. Similarly, $a''=0$, $a''=1$ and $a''=2$ refer to the actions of increasing, decreasing or keeping the current price, respectively. Intuitively, this definition of the action space implies a total of 9 different combination of actions, i.e., $(3\times3)$ actions, at each time slot. This strategy significantly reduces the action space size which helps the DRL to converge quickly.

\textit{Immediate Reward:}
Since our objective in the contract is to maximize~\eqref{eq:optz_UppLinkLayer}, we craft the immediate reward function to align with this objective.
To incorporate the IC and IR constraints in~\eqref{eq:optz_UppLinkLayer} into the immediate reward function, we design a multi-objective reward function based on weighted sum technique.
Specifically, we define the reward function as follows:

\begin{equation}\label{eq:immediate_reward_UplinkLayer}
\begin{split}
r^{\dagger}(\mathcal{S}^{\dagger(t)}, a^{\dagger(t)}, \mathcal{S}^{\dagger(t+1)}) = w_1 \sum\limits_{\lambda\in\Lambda}\sum\limits_{\gamma\in\Gamma}\sum\limits_{\psi\in\Psi}n_{\lambda,\gamma,\psi}\Bigl(
        \sigma f(\hat{s}^{(t)}_{\lambda,\gamma,\psi})
        - \pi^{\dagger,(t)}_{\lambda,\gamma,\psi}
        +(K - \varpi_{\gamma})\Bigr) \\
 + w_2\big[\sum\mathbf{x}^{\dagger(t)} - \sum\mathbf{x}^{\dagger(t+1)}\big] + w_3\big[\sum\mathbf{y}^{\dagger(t)} - \sum\mathbf{y}^{\dagger(t+1)}\big],
\end{split}
\end{equation}

\noindent where $w_1 + w_2 +w_3 =1$ are the weight factors of each term in~\eqref{eq:immediate_reward_UplinkLayer} and $n_{\lambda,\gamma,\psi}$ is the number of sensing IoT devices with type-$(\lambda,\gamma,\psi)$.
The first term in~\eqref{eq:immediate_reward_UplinkLayer} reflects the objective of maximizing the VSP's revenue. The second and third terms reflect the objective of reducing the number of violations of IR and IC properties, respectively.
Note here that rewards are only received after the increment of the timestep.

\textit{Optimization Formulation:}
The objective is to find a policy $\textbf{p}^{\dagger*}$ that has the best mapping from states to actions which maximizes the average long-term reward $\mathcal{R}(\textbf{p}^{\dagger})$.
Formally, the optimization problem is defined as
\begin{equation}\label{eq:DRL_Obj_UpstreamLayer}
    \max\limits_{\textbf{p}^{\dagger}} \qquad
    \mathcal{R}(\textbf{p}^{\dagger}) = \lim\limits_{\Upsilon\rightarrow\infty}\frac{1}{\Upsilon}\sum\limits_{t=1}^{\Upsilon}\mathbb{E}(r^{\dagger}_{t}(s^{\dagger}_t, \textbf{p}^{\dagger}(s^{\dagger}_t))),
\end{equation}

\noindent where $r^{\dagger}_{t}(s^{\dagger}_t, \textbf{p}^{\dagger}(s^{\dagger}_t))$ is the immediate reward under policy $\textbf{p}^{\dagger}$ at time $t$ defined in~\eqref{eq:immediate_reward_UplinkLayer}\footnote{Note that the unique ability of our solution is at optimizing for other objectives. Specifically, we might have other objectives that can be simply achieved by modification to the reward function, e.g., maximizing the social welfare of the system.}.


The standard approach to solve the MDP described earlier is to adopt one of the available single-agent DRL algorithms, e.g., Deep Q-Network (DQN) or Proximal Policy Optimization (PPO)~\cite{Sutton2018}.
However, standard single agent DRL algorithms cannot solve the described MDP. Specifically, in single agent DRL and at each time slot, the agent extracts a single action to perform. However, in our MDP there is a need to perform $N$ actions simultaneously. 
As there are several actions to be executed simultaneously, an attractive approach is to adopt multi-agent reinforcement learning (MARL)~\cite{Sutton2018}. In MARL systems, several agents are trained to work independently to achieve one goal or compete against each other. However, our studied system also differs from these settings as we only want to train the VSP to derive the optimal contract and there are no other agents to train. Inspired by MARL and the work in~\cite{Hongzi_HotNet_2016}, we develop a novel MARL architecture to solve the optimal iterative contract problem.
In what follows, we first continue the formulation of the downstream layer and its corresponding MDP. Next, we describe the details of our proposed learning-based iterative contract.

\subsection{Downstream Layer (VSP and Metaverse users)}

In this layer, the objective of the VSP is to find a set of qualities of the delivered digital twin jointly with their respective prices to maximize its revenue.
As earlier described, the quality of the delivered digital twin is measured using the resolution (which reflects the perception) and refresh rate (which reflects the timeliness of the information).
Hereafter, we denote the set of available resolutions as $\mathcal{R}$, the set of refresh rates as $\mathcal{H}$, and the set of prices as $\Pi^{\ddagger}$.
Here we consider that the different combinations of resolutions and refresh rates are referred to by an auxiliary variable $q$.
For each Metaverse user with type-($\tau, \phi$), the VSP assigns a quality $q_{\tau, \phi}$ and charges a price $\pi^{\ddagger}_{\tau, \phi}$. The set of quality-price combinations is denoted as $\Omega^{\ddagger} = \{(q_{\tau, \phi}, \pi^{\ddagger}_{\tau, \phi}) |\forall \tau \in \Xi, \forall \phi \in \Phi \}$.
The IR and IC properties of the downstream layer are defined as follows.

\begin{definition}
\textit{Individual Rationality (IR) for Metaverse user: A Metaverse user with type-($\tau, \phi$) will only accept to purchase the digital twin from the VSP if its utility\footnote{Note that we use $V^{\ddagger}(\tau, \phi, q_{\tau, \phi})$ instead of $V^{\ddagger}(\tau, \phi, r, h)$ for notational consistency.} is non-negative, i.e.,}
\begin{equation}\label{eq:IR_MV}
     V^{\ddagger}(\tau, \phi, q_{\tau, \phi})  - \pi^{\ddagger}_{\tau, \phi} \geq 0, \quad \forall \tau \in \Xi, \forall \phi \in \Phi.
\end{equation}
\end{definition}

\begin{definition}
\textit{Incentive Compatibility (IC) for Metaverse user:
The utility of a Metaverse user with type-($\tau, \phi$) is maximized only when selecting the contract designed for its true type, i.e.,
}
\begin{equation}\label{eq:IC_MV}
\begin{split}
     V^{\ddagger}(\tau, \phi, q_{\tau, \phi})  - \pi^{\ddagger}_{\tau, \phi} \geq V^{\ddagger}(\tau, \phi, q_{\tau', \phi'})  - \pi^{\ddagger}_{\tau', \phi'},\\
     \quad \forall \tau,\tau' \in \Xi, \forall \phi,\phi' \in \Phi, \tau' \neq \tau, \phi' \neq \phi.
\end{split}
\end{equation}
\end{definition}

Since the VSP dominates the trading process, we model the digital twin trading as a \textit{monopoly market}, in which the VSP's objective is to maximize its overall utility, which is written as 
\begin{equation}
        R^{\ddagger}(\Omega^{\ddagger}) = \sum\limits_{\tau\in\Xi}\sum\limits_{\phi\in\Phi} M Q^{\ddagger}(\tau,\phi) \left(\pi^{\ddagger}_{\tau,\phi} - \Upsilon^{\ddagger}(q_{\tau,\phi})\right),
\end{equation}

\noindent where $Q^{\ddagger}(\tau,\phi)$ is the joint probability mass function of the Metaverse users having type-($\tau, \phi$) and is obtained from previous observations~\cite{Zehui_2020_TMC}.
For instance, each Metaverse user device support a different frame rate and have different valuation towards them, which is a private information for each Metaverse user. Nevertheless, the VSP has some prior knowledge about their probability distribution which is modeled here using $Q^{\ddagger}(\tau,\phi)$.
In order for the contract to be feasible, it has to guarantee both IC and IR. Therefore, the optimal contract can be derived by solving the following problem:
\begin{subequations}
\label{eq:optz_DownlinkLayer}
\begin{align}
\begin{split}
\mathcal{P}_2: \max\limits_{\boldsymbol{(q^{\ddagger}, \pi^{\ddagger})}} \sum\limits_{\tau\in\Xi}\sum\limits_{\phi\in\Phi} M Q^{\ddagger}(\tau,\phi) \left(\pi^{\ddagger}_{\tau,\phi} - \Upsilon^{\ddagger}(q_{\tau,\phi})\right),
\end{split}\\
\begin{split}
\hspace{1cm} s.t. \quad \eqref{eq:IR_MV} ~and~\eqref{eq:IC_MV}. \label{eq:IR_IC_MV}
\end{split}
\end{align}
\end{subequations}

However, the parameters in \eqref{eq:IR_MV} and~\eqref{eq:IC_MV} are private to the Metaverse users and can be misreported. Similar to the upstream layer problem, this problem is addressed using DRL. Therefore, in what follows, we first start by describing the MDP of the downstream layer. Next, the solution of this MDP and that of the upstream layer is described in detail.

\textbf{Markov Decision Process:}
The MDP of the downstream layer is defined by the tuple $<~\mathcal{S}^{\ddagger}, \mathcal{A}^{\ddagger}, \mathcal{P}^{\ddagger}, r^{\ddagger}>$ where $\mathcal{S}^{\ddagger}$ is the state space, $\mathcal{A}^{\ddagger}$ is the action space, $\mathcal{P}^{\ddagger}$ is the state transition probabilities and $r^{\ddagger}$ is the immediate reward received by the agent, i.e., the VSP, after performing action $a^{\ddagger}$ at state $s^{\ddagger}$.

\textit{State Space:}
The state space of the system at time slot $t~(t=1,2,\dots,T)$ is defined as
\begin{equation}\label{eq:state_space_DownLinkLayer}
\begin{split}
\mathcal {S}^{\ddagger(t)} \triangleq \Big \{ \mathbf{a}^{\ddagger(t-1)}, \boldsymbol{\pi}^{\ddagger(t)}, \boldsymbol{q}^{(t)},
\mathbf{x}^{\ddagger(t)}, \mathbf{y}^{\ddagger(t)} \Big \},
\end{split}
\end{equation}

\noindent where $\mathbf{a}^{\ddagger(t-1)}$ is the action vector from the previous time slot, $\boldsymbol{\pi}^{\ddagger(t)}$ is the price vector at time slot $t$ and $\boldsymbol{q}^{(t)}$ is the digital twin quality vector at time slot $t$. $\mathbf{x}^{\ddagger(t)}$ and $\mathbf{y}^{\ddagger(t)}$ are binary vectors of Metaverse users which have their IR an IC violated, respectively.
The system state is then defined as a composite variable $\mathbf{s}^{\ddagger} \triangleq (\mathbf{a}^{\ddagger}, \boldsymbol{\pi}^{\ddagger}, \boldsymbol{q},
\mathbf{x}^{\ddagger}, \mathbf{y}^{\ddagger}) \in \mathcal{S}^{\ddagger}$.

\textit{Action Space:}
The action space for the downstream layer is identical to that of the upstream layer with difference only in adjusting the quality instead of the semantic information size.

\textit{Immediate Reward:}
The reward function of the downstream layer is crafted to maximize~\eqref{eq:optz_DownlinkLayer} while incorporating the IR and IC constraints defined in~\eqref{eq:IR_MV} and~\eqref{eq:IC_MV}, respectively. Therefore, the reward function of the downstream layer is formalized as
\begin{equation}\label{eq:immediate_reward_DownLinkLayer}
\begin{split}
r^{\ddagger}(\mathcal{S}^{\ddagger(t)}, a^{\ddagger(t)}, \mathcal{S}^{\ddagger(t+1)}) = w'_1 \sum\limits_{\tau\in\Xi}\sum\limits_{\phi\in\Phi}n_{\tau,\phi}\Bigl(
        \pi^{\ddagger,(t)}_{\tau,\phi}
        - \Upsilon^{\ddagger}(q_{\tau,\phi}) \Bigr) 
 + w'_2\big[\sum\mathbf{x}^{\ddagger(t)} - \sum\mathbf{x}^{\ddagger(t+1)}\big] \\+ w'_3\big[\sum\mathbf{y}^{\ddagger(t)} - \sum\mathbf{y}^{\ddagger(t+1)}\big],
\end{split}
\end{equation}

\noindent where $w'_1 + w'_2 +w'_3 =1$ are the weight factors and $n_{\tau,\phi}$ is the number of Metaverse users with type-$(\tau,\phi)$.

\textit{Optimization Formulation:}
The optimization problem of the downstream layer is defined as
\begin{equation}\label{eq:DRL_Obj_DownstreamLayer}
    \max\limits_{\textbf{p}^{\ddagger}} \qquad
    \mathcal{R}(\textbf{p}^{\ddagger}) = \lim\limits_{\Upsilon\rightarrow\infty}\frac{1}{\Upsilon}\sum\limits_{t=1}^{\Upsilon}\mathbb{E}(r^{\ddagger}_{t}(s^{\ddagger}_t, \textbf{p}^{\ddagger}(s^{\ddagger}_t))),
\end{equation}

\noindent where $r^{\ddagger}_{t}(s^{\ddagger}_t, \textbf{p}^{\ddagger}(s^{\ddagger}_t))$ is the immediate reward under policy $\textbf{p}^{\ddagger}$ at time $t$ defined in~\eqref{eq:immediate_reward_DownLinkLayer}.

\subsection{Iterative Contract Design}
\begin{figure}[ht!]
    \centering
    \includegraphics[width=.45\textwidth,height=4.5cm]{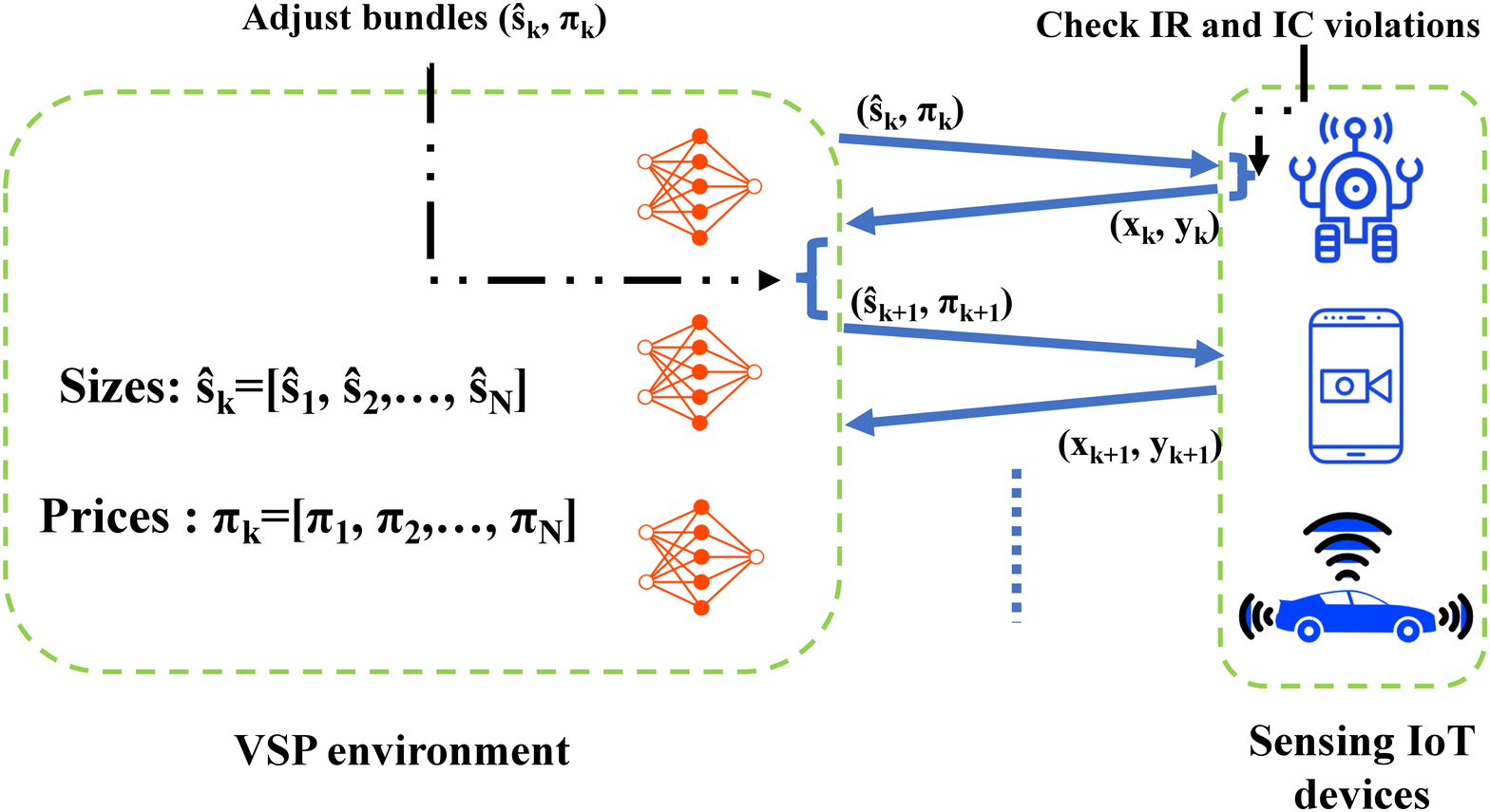}
    \caption{Proposed learning-based iterative contract.}
    \label{fig:MARL_Description}
\end{figure}

The proposed learning-based iterative contract is shown in Fig.~\ref{fig:MARL_Description} while Algorithm~\ref{algo:MARL_PDDQL} summarizes the major steps. Here we describe the framework with respect to the upstream layer while its application on the downstream layer is straightforward as clearly seen from the similarity between their MDPs.

Specifically, the algorithm (administered by the VSP) starts by initializing the semantic information size vector $\boldsymbol{\hat{s}}$ and the price vector $\boldsymbol{\pi^{\dagger}}$ based on a uniform distribution from the intervals $[size_k \times (1-range), size_k \times (1+range)]$ and $[price_k \times (1-range), price_k \times (1+range)]$, respectively.
In addition, the binary vectors $\mathbf{x}^{(t)}$ and $\mathbf{y}^{(t)}$ are initially set to 1 for all the vectors' elements.
Next, the VSP initializes $N$ single-agent DRL networks to learn the optimal strategy for adjusting the bundles of each sensing IoT device. As there are a variety of DRL algorithms with some algorithms working better in some domains than others, we adopt our previously developed prioritized double deep Q-Learning (PDDQL) algorithm in~\cite{Ismail_2021_TVT} and refer the reader for the detailed description therein.
At the very first iteration of the algorithm, the VSP populates the initial set of bundles directly to the $N$ sensing IoT devices. The number of different types in the multi-dimensional contract is extracted from previous interactions with the sensing IoT devices. Once the sensing IoT devices receive the contract bundles, each sensing IoT device verifies whether its IR or IC is violated based on~\eqref{eq:IR_IoT} and~\eqref{eq:IC_IoT} and then returns a binary tuple $(x, y)$ to the VSP. The VSP then constructs the full vectors $\mathbf{x}^{(t)}$ and $\mathbf{y}^{(t)}$ and shares their content with each agent in its environment. We refer to this step as augmentation of the MDP as the agents are augmented with information initially unobservable about the states of each other (i.e., IR and IC violations, previous actions, semantic information sizes and prices). At this point, each agent executes the PDDQL algorithm to extract the optimal adjustment to be performed based on the set of actions as earlier defined. 
As such, we name this algorithm augmented multi-agent PDDQL (MA-PDDQL).
Next, the VSP performs the appropriate adjustments for each bundle, i.e., semantic information sizes and prices, and then delivers the new set of bundles to the sensing IoT devices to start the next round.
Once we reach a state where $x_i^{(t)} = y_i^{(t)} = 0$ for all the vectors elements, the derived contract is considered feasible and satisfy IR and IC conditions. However, the solution is not necessarily optimal. Several round needs to be executed until no improvement in the VSP's utility/revenue is obtained. For this reason, we call our framework as an \emph{interactive contract} where the optimal contract is derived based on several rounds of interaction between the VSP and the sensing IoT devices.

We should note here some key features in the design of our proposed framework:
\begin{itemize}
    \item First, a technical challenge to solve is the convergence of the DRL-based contract as the prices of all bundles change simultaneously. Specifically, the very frequent changes in the price or the semantic information size of one bundle affects the strategy of all the participants when choosing their optimal contract bundle.
    This makes the DRL environment non-stationary and noisy for each agent to learn a stable policy. 
    We address this issue by establishing a virtual communication channel between the learning agents in the MARL environment, which we refer to as augmentation of the MDP.
    Specifically, after receiving the IR and IC tuples from all the sensing IoT devices, the full vectors $\mathbf{x}^{(t)}$ and $\mathbf{y}^{(t)}$ are created and shared as part of the state of each agent. In addition, each agent in the VSP environment is aware of the current set of bundles and the previously taken actions by all other agents. This augmentation of the observation space for each agent makes the MDP easily learnable and the agents can then learn from collective experiences.

    \item In traditional contracts, the distribution of the types is assumed to be known and is necessary to drive the optimal set of bundles (e.g., see~\cite{Gao_2011, Xiong_2020_TWC_Contract}). However, if the distribution changes, the already derived solution will not be optimal and might become infeasible (IR and IC will be violated). In our learning-based contract, there is no need to have this prior information. 
        
    \item The MA-PDDQL algorithm requires from the sensing IoT devices only a flag about their IR and IC status and not their private types, which is totally different from existing contract solutions that requires the disclosure of these information (referred to as the revelation principal in contract theory~\cite{Bolton_2005}). Some privacy-sensitive participants may be reluctant about engaging in such contracts as their private information might be used for other purposes beyond the contract, e.g., delivering dedicated advertisement as studied in~\cite{Boyang_INFOCOMW_2018}. Our design preserves the privacy of the participants about their private types which is a major usefulness of our proposed framework.

    \item A major issue to address is the willingness towards untruthful behavior by the sensing IoT devices during the learning process, which leads to the violation of the IR and IC properties. In our framework, the participants have no information about their utilities in the next round as the VSP will adjust all the bundles in an unpredictable strategy. This can cause the utility of a participant to decrease in the current state, which can be the final state, compared to a previous state where its utility was higher. Therefore, the participants are incentivized not to misreport their true state, i.e., IR and IC violations.
\end{itemize}

\begin{algorithm}[ht!]
\SetAlgoLined
\SetKwInOut{Input}{Input}
\SetKwInOut{Output}{Output}
 \Input{Initialize semantic information size and prices to random values.}
 \Output{Optimal semantic information sizes and prices $\boldsymbol{(\hat{s}, \pi^{\dagger})}$.}
 
 \For{$t=1,2, \ldots$ \KwTo $convergence$}{
    Initialize empty action list\;
     \For{$i=1,2, \ldots$ \KwTo $N$}{
        Select action for device $i$ based on PDDQL and append to the action list\;
     }
    Execute the simultaneous actions from the action list and get the next state\;
    \For{$i=1,2, \ldots$ \KwTo $N$}{
        Store the tuple $(s_t, a_t, s_{t+1}, r_t)$ and update the policy of agent $i$\;
     }  
}
 \caption{Augmented MA-PDDQL Algorithm pseudo-code}
 \label{algo:MARL_PDDQL}
\end{algorithm}

Next, we evaluate the proposed learning-based iterative multi-dimensional contract framework.

\section{Numerical Evaluation}\label{section_Results}
In this section, we validate the performance of our proposed iterative contract for the Metaverse through extensive simulations.
As in~\cite{Ismail_2022_FNWF}, we use existing semantic extraction algorithms to process images and radar signals to extract the semantics~\cite{He_2017_ICCV, Ouaknine_2021_ICCV}. 
As stated before, the structure of our iterative contract in the upstream layer is quite similar to the one in the downstream layer. Therefore, we present here the numerical results for the upstream layer only.

\subsection{Simulation Settings}
Unless otherwise stated, the DRL algorithm is trained over 700 episode with 200 iterations on each episode.
In addition, the weighting factors in the reward function are all set to $0.33$. We also consider a total of 27 (3x3x3) different sensing IoT device types.
The type of each sensing IoT device is chosen uniformly from the set of possible joint types $(B\times C\times E)$.
Furthermore, the types and other variables such as the transmit power and the energy consumption of the sensing IoT devices are normalized between 0 and 1.  
We consider that $range=0.9$ while $n_{1,k}$ and $n_{2,k}$ take values from the discrete set $[-0.9, -0.7,\ldots,0.7,0.9]$.

\subsection{Benchmarking Scheme}
Due to the novelty of our developed MA-PDDQL and the iterative contract design, it is difficult to find existing baselines to compare with. Therefore, to evaluate and show the benefits of our novel augmented MA-PDDQL algorithm, we design a baseline scheme called Naive MA-PDDQL based on~\cite{Gronauer_2021}. Different from the augmented MA-PDDQL, the naive MA-PDDQL uses a partially observed MDP (POMDP) for each agent. Specifically, the POMDP state is defined as

\begin{equation}\label{eq:state_space_UplinkLayer_Naive}
\begin{split}
\mathcal {S}^{\dagger(t)}_{naive} \triangleq \Big \{ a^{\dagger(t-1)}, \pi^{\dagger(t)}, \hat{s}^{(t)},
x^{\dagger(t)}, y^{\dagger(t)} \Big \},
\end{split}
\end{equation}

The action set and immediate reward function are also scaled down to the case of single observations. Since the state observation of each agent does not represent the whole system state as earlier defined in the augmented MDP, each agent has only a partial observation. 

\begin{figure*}[!h] 
     \centering
     \begin{subfigure}[b]{0.3\textwidth}
         \centering
         \includegraphics[width=\textwidth]{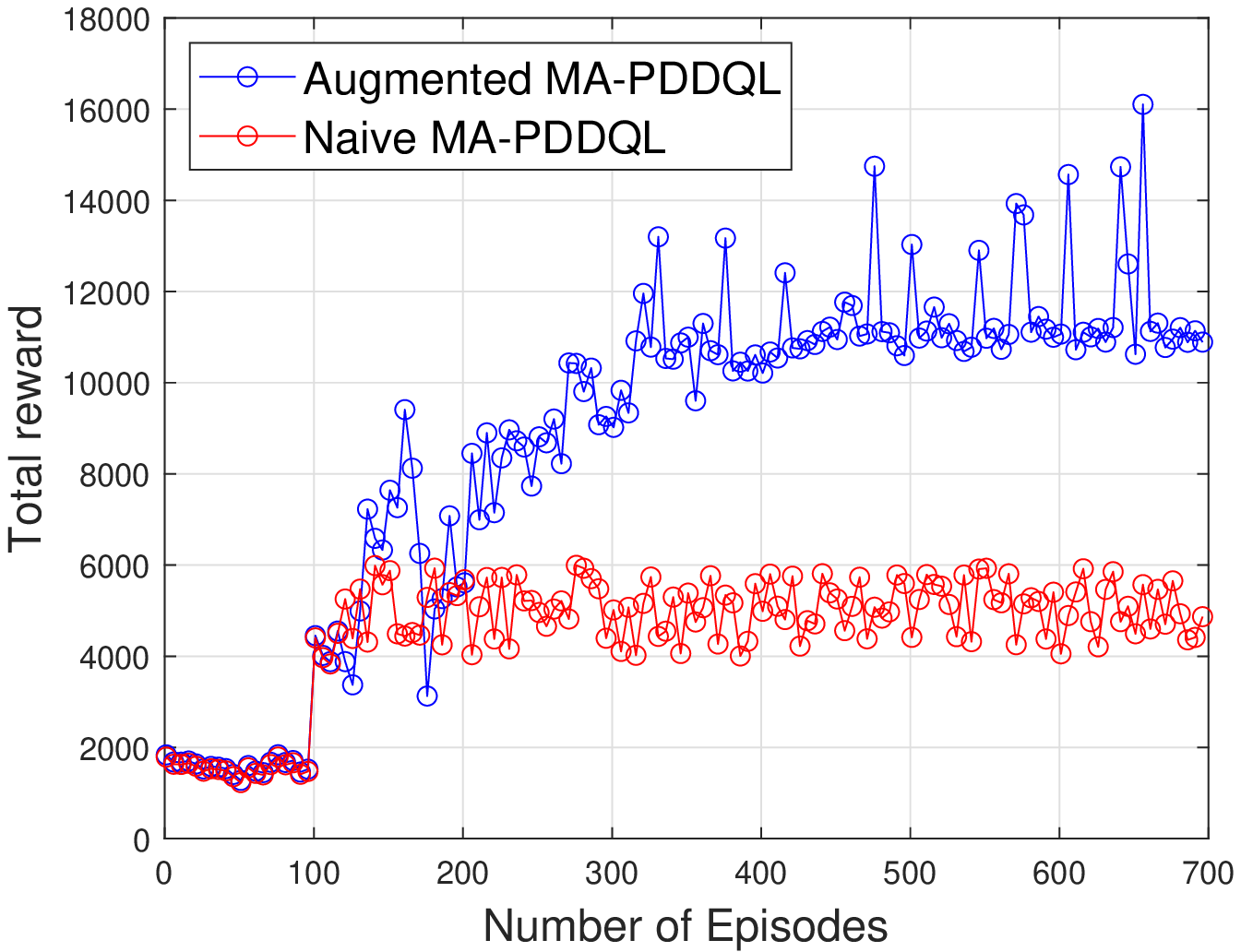}
         \caption{}
         \label{fig:Learning_rewards}
     \end{subfigure}
     \hfill
     \begin{subfigure}[b]{0.3\textwidth}
         \centering
         \includegraphics[width=\textwidth]{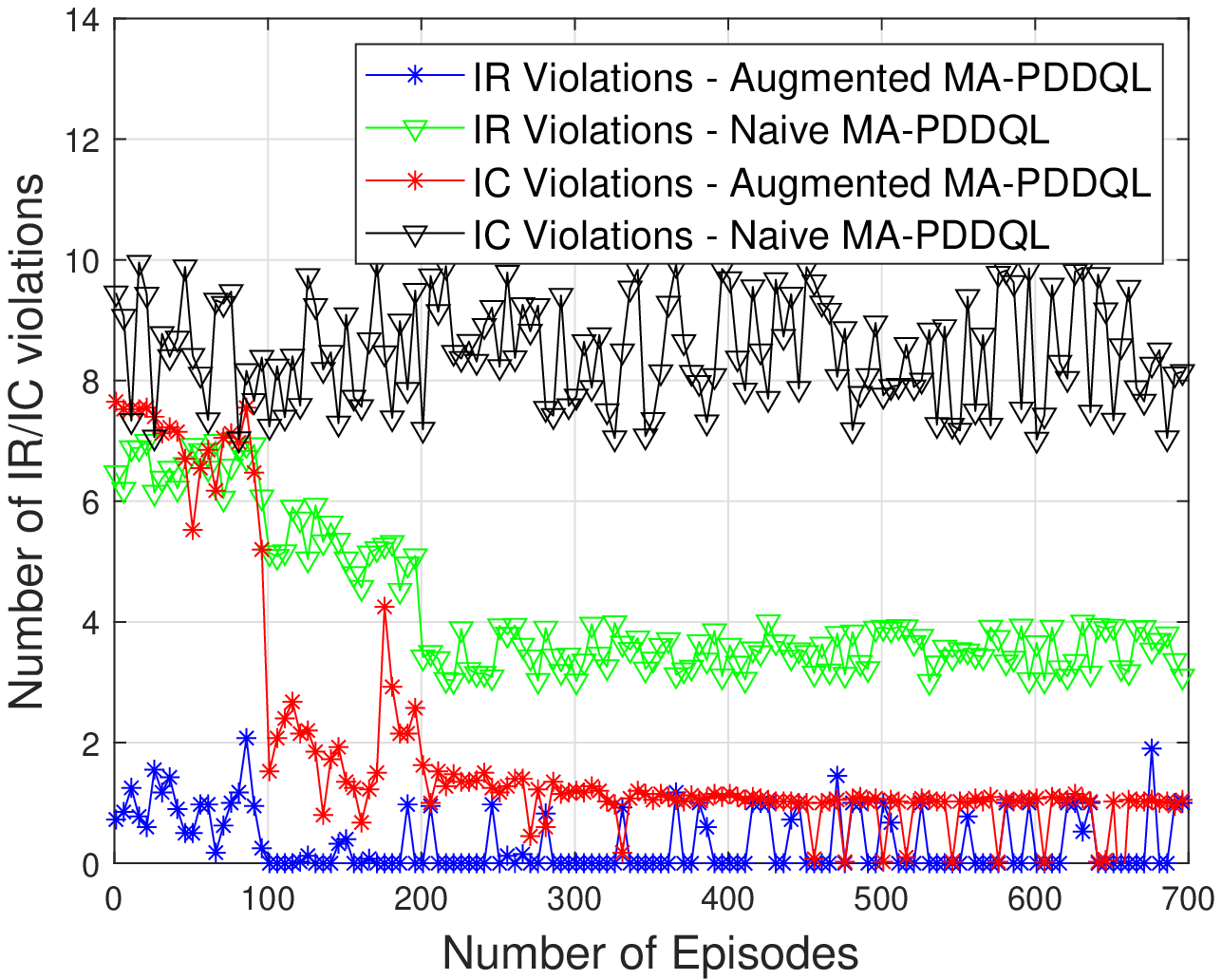}
         \caption{}
         \label{fig:Learning_IR_IC_Violations}
     \end{subfigure}
     \hfill
     \begin{subfigure}[b]{0.3\textwidth}
         \centering
         \includegraphics[width=\textwidth]{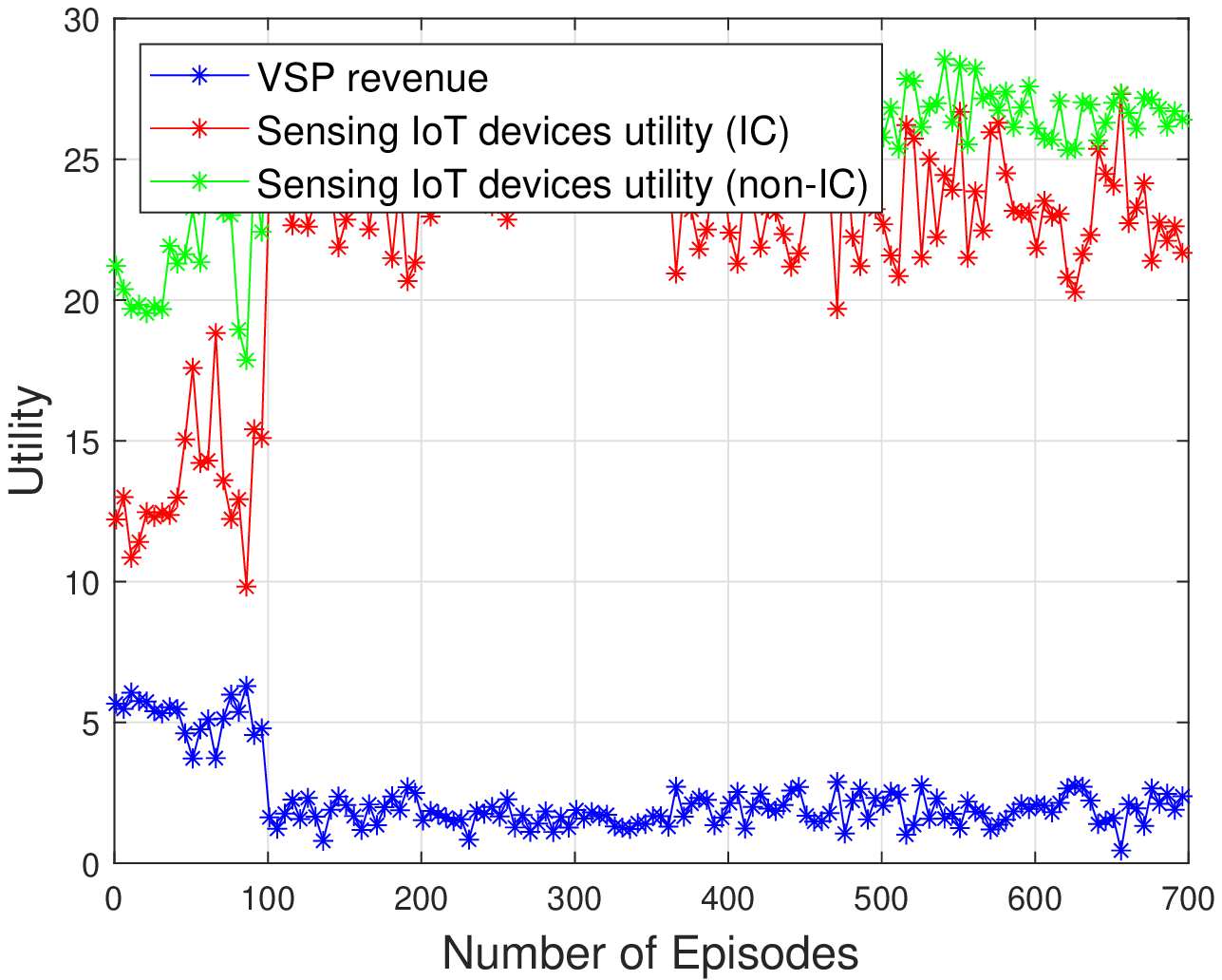}
         \caption{}
         \label{fig:Learning_VSP_revenue}
     \end{subfigure}
        \caption{(a) Total reward for each episode. (b) Average number of IR and IC violations. (c) Average revenue of the VSP and sensing IoT devices.}
        \label{fig:Learning_progress}
\end{figure*}
\subsection{Results}

\subsubsection{Convergence analysis and validity of the feasibility conditions}

To observe the convergence behavior of our learning-based iterative contract, we measure the number of IR and IC violations and the revenue of the VSP at the last iteration of each episode.
We observe from Fig.~\ref{fig:Learning_rewards} that the average reward of the augmented MA-PDDQL stabilizes after 350 episodes. However, the average reward of the naive MA-PDDQL is lower than that of the augmented MA-PDDQL and stops increasing after 100 episode only, indicating that the naive MA-PDDQL is not able to converge. 
As observed from Fig.~\ref{fig:Learning_IR_IC_Violations}, the number of IR and IC violations for the augmented MA-PDDQL decreases as the algorithm progress in learning.
Interestingly, we observe from Fig.~\ref{fig:Learning_IR_IC_Violations} that there is an improvement in the minimization of the IR violations for the naive MA-PDDQL while no significant change occurs for the number of IC violations. This is justified by the fact that the IR constraint is much simpler than the IC constraint. The IR constraint needs only to guarantee that the utility of the participants is non-negative, while the IC constraint needs to guarantee that the utility of a participant is maximized for the true type of the participant compared to all other types. The latter cannot be learned by the naive MA-PDDQL because of the non-stationarity problem of the POMDP. Specifically, as each agent in the naive MA-PDDQL environment observes only its private state, and thus is unaware of other agents changes of their bundles, it is unable to find an optimal adjustment for its bundle to meet the IC constraint for its respective sensing IoT device. 


To dive further in the structure of our learning-based algorithm, we plot in Fig.~\ref{fig:Learning_VSP_revenue} the average revenue of the VSP as the training progress. Remarkably, we observe that the average revenue of the VSP decreases as the training progress, which seems to behave against our main objective, i.e., maximization of revenue of the VSP as stated in $\mathcal{P}_1$. However, we should understand from Fig.~\ref{fig:Learning_IR_IC_Violations} that in the first few episodes, the obtained revenue of the VSP is achieved while having the IR and IC properties violated for the majority of the participants, i.e., sensing IoT devices. This implies that the majority of the sensing IoT devices will behave untruthfully and select contract items not dedicated for their true types, making the realized utility of the VSP very low compared to the expected one. At the end of the training, the derived VSP utility is achieved with majority of the IR and IC satisfied.
Here, we should note that an important difference between our learning-based contract and existing works on contract theory is the satisfaction of the feasibility conditions, i.e., IR and IC, under information asymmetry. In classical approaches, the IR and IC constraints need to be satisfied for all the participants, i.e., sensing IoT devices. However, in our framework we aim towards the minimization of their occurrence.

We also plot how the utility of the sensing IoT devices changes as the training progresses in Fig.~\ref{fig:Learning_VSP_revenue}. The red color refer to the utility of the sensing IoT devices in the case that they have chosen their designated bundle (i.e., IC preserved). The green color, however, refers to the case when the sensing IoT devices choose the bundle that maximize their utility.
In both scenarios, we observe that the utility of the sensing IoT devices increases then stabilize after episode 100 while the revenue of the VSP decreases and then stabilizes, which is counter-intuitive. To explain this behavior, we first note that based on the objective function in $\mathcal{P}_1$, we should only maximize the revenue of the VSP.
From~\eqref{eq:IR_IoT}, it seems that the optimal strategy for the VSP is to set the price for each contract item equal to their valuation by their corresponding sensing IoT devices. In this case, the utility of the sensing IoT devices will be equal to zero. Based on this observation, we expect the revenue of the VSP to increase and the utilities of the participants to decrease. 
However, as the objective function of problem $\mathcal{P}_1$ is adjusted in the reward function of the MDP, an action that further minimizes or maintains the number of IR and IC violations is given a positive reward (see the last term in~\eqref{eq:immediate_reward_UplinkLayer}). 
Therefore, the derived bundles are not pushed towards minimizing the gap between the provided prices and the private valuations of each sensing IoT device, which justifies the increase of the utility of the sensing IoT devices.

\begin{figure*}[!h] 
     \centering
     \begin{subfigure}[b]{0.3\textwidth}
         \centering
         \includegraphics[width=\textwidth]{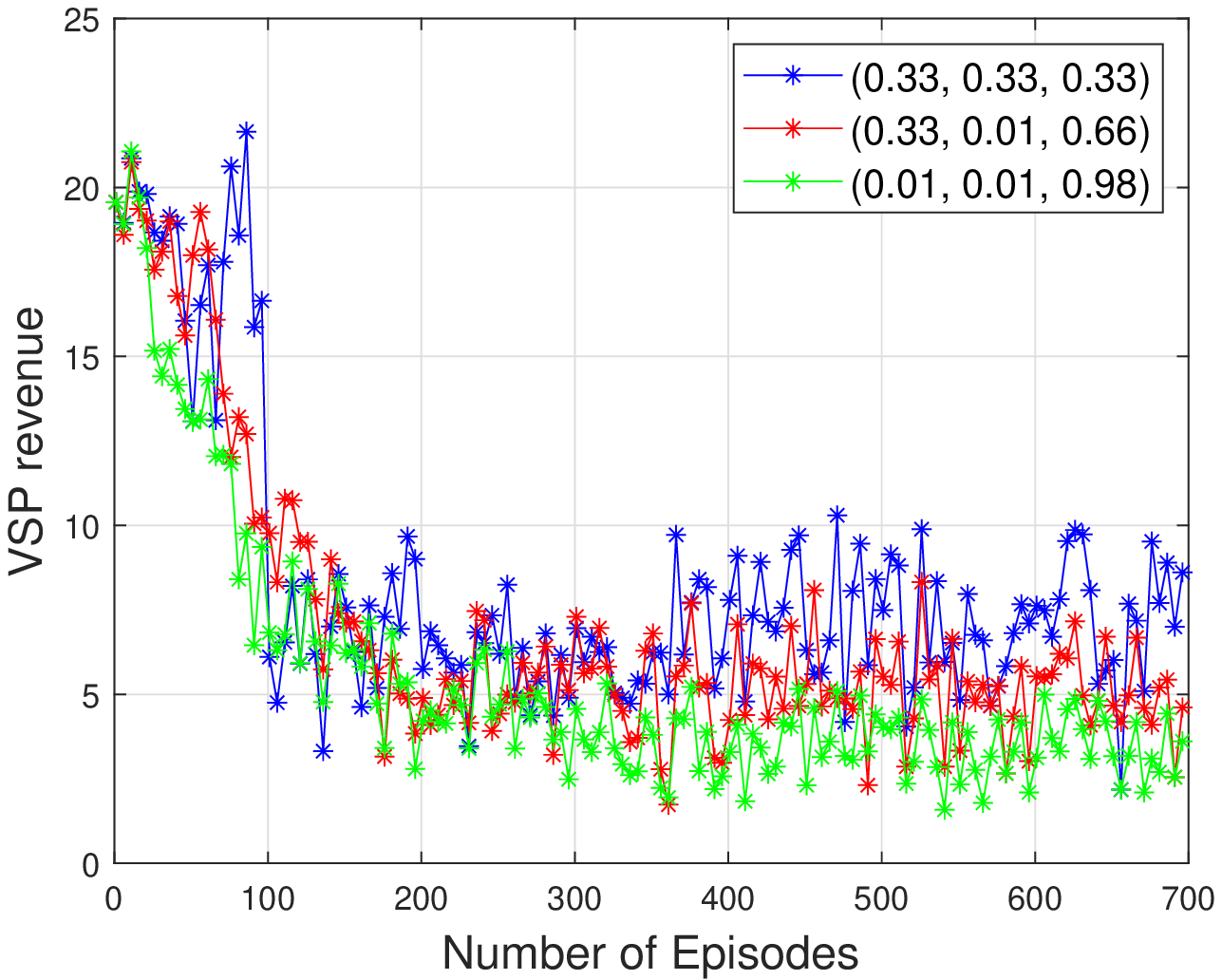}
         \caption{}
         \label{fig:weighting_factors_VSP_revenue}
     \end{subfigure}
     \hfill
     \begin{subfigure}[b]{0.3\textwidth}
         \centering
         \includegraphics[width=\textwidth]{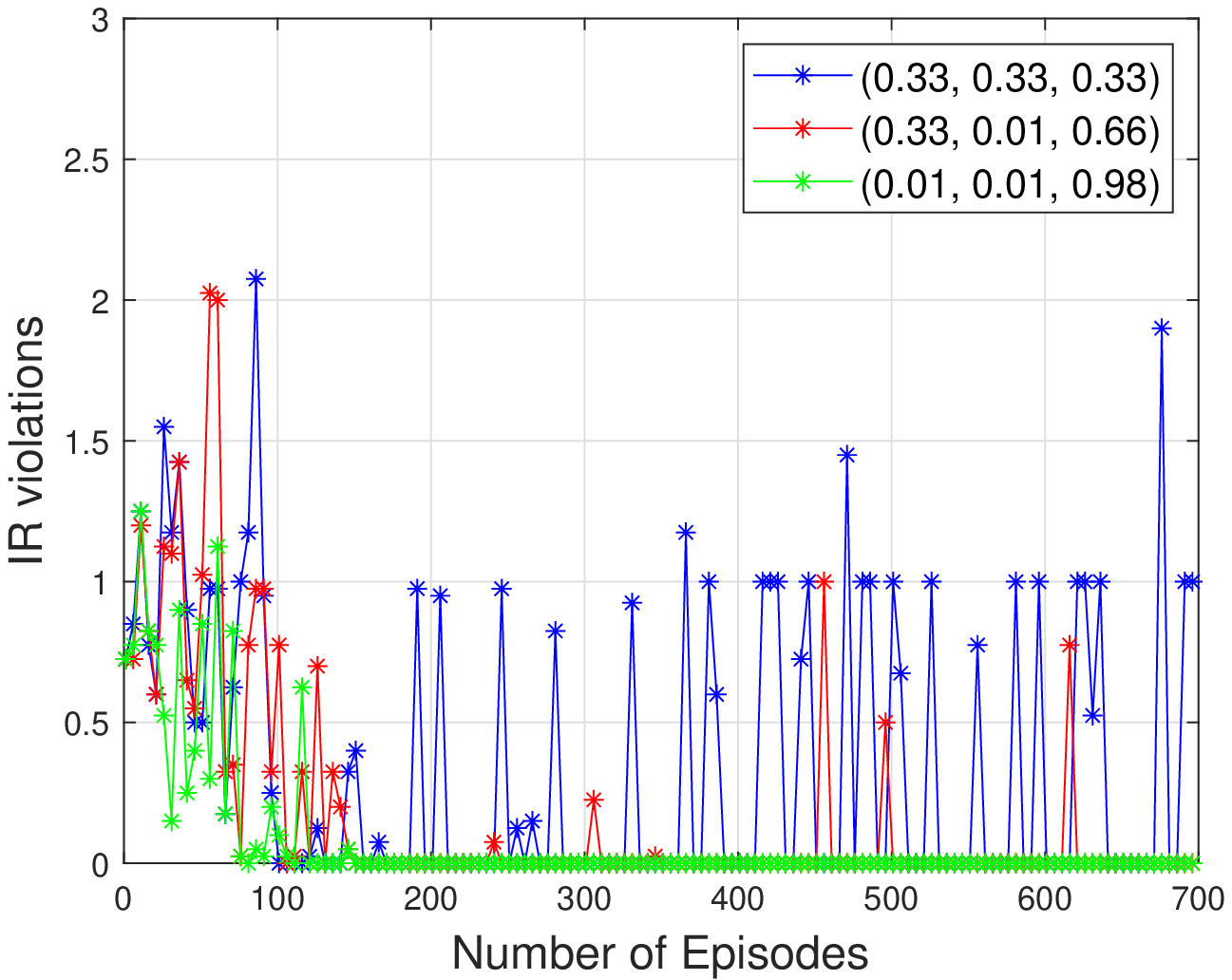}
         \caption{}
        \label{fig:weighting_factors_IR_Violations}
     \end{subfigure}
     \hfill
     \begin{subfigure}[b]{0.3\textwidth}
         \centering
         \includegraphics[width=\textwidth]{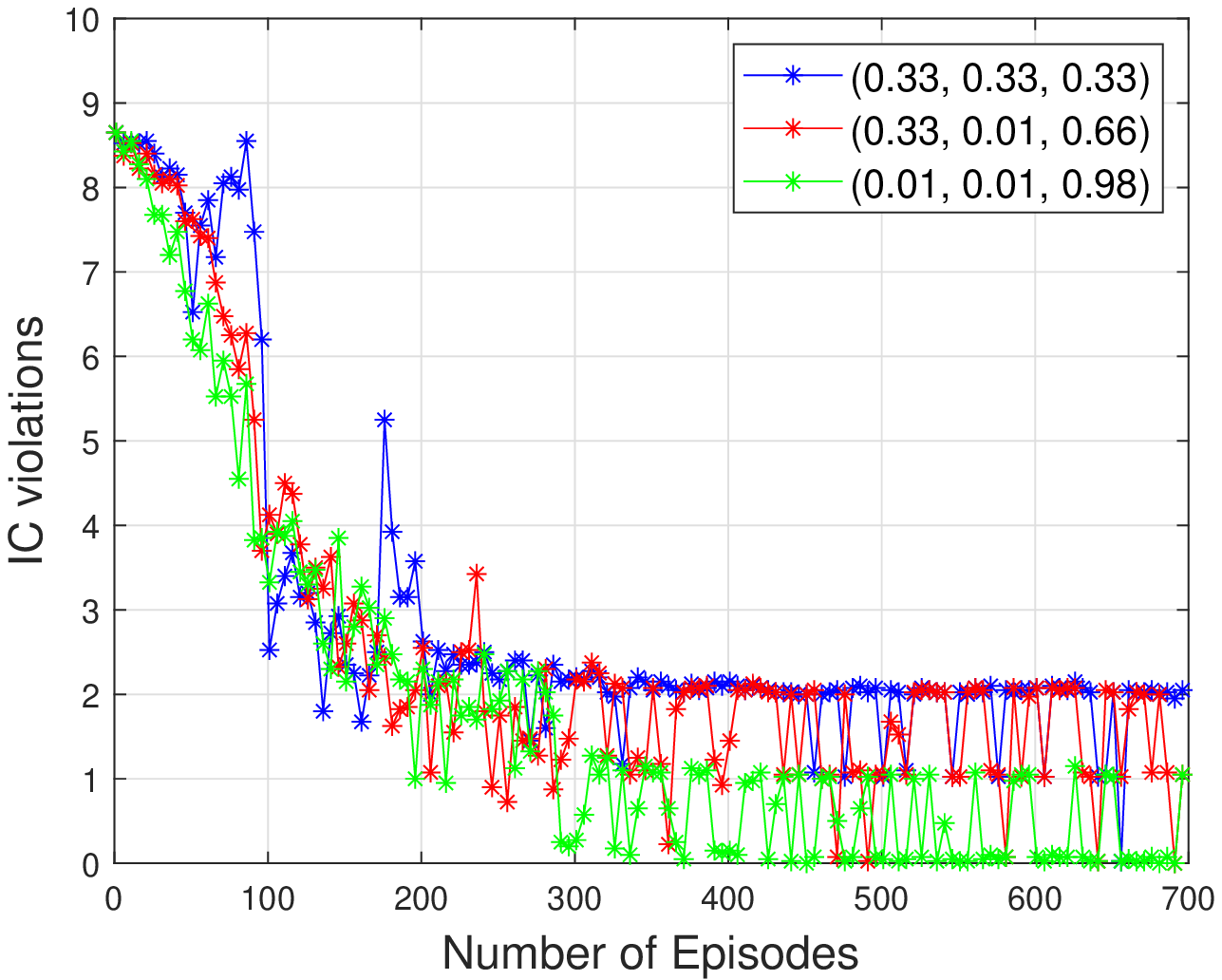}
         \caption{}
         \label{fig:weighting_factors_IC_Violations}
     \end{subfigure}
        \caption{Impact of the weighting factors: (a) average VSP revenue. (b) and (c) average number of IR and IC violations.}
        \label{fig:weighting_factors}
\end{figure*}

\subsubsection{Impact of the weighting factors}
Motivated by the previous results from Fig.~\ref{fig:Learning_VSP_revenue}, we further study the impact of the weighting factors in the reward function on the performance of our framework. In this experiment, we set three different scenarios for the values of $w_1$, $w_2$ and $w_3$ and observe how the VSP revenue and the IR and IC violations changes.
Specifically, we consider the following scenarios: $(w_1,w_2,w_3)=(0.33,0.33,0.33)$, $(w_1,w_2,w_3)=(0.33,0.01,0.66)$ and $(w_1,w_2,w_3)=(0.01,0.01,0.98)$.
The results are shown in Fig.~\ref{fig:weighting_factors}.
Interestingly, we observe from Fig.~\ref{fig:weighting_factors_IR_Violations} that putting a weight of $0.01$ on the IR term gives better results for the IR violation compared to the case of setting a higher weight $0.33$. This is explained by the fact that IR is satisfied for the majority of the sensing IoT devices, which is dependant on the sets of semantic information size and prices that the MA-PDDQL is learning on. 
We should also note that putting more weight on the IC term helps reducing the IR violation further and hence, minimizes the influence of the weight term of the IR term. Specifically, during the IC property verification, each sensing IoT device compares its utility when choosing its true type with the case of choosing any other type. Therefore, if its IC property is not violated, its utility is unlikely to be negative.

\begin{figure}
     \centering
     \begin{subfigure}[b]{0.3\textwidth}
         \centering
         \includegraphics[width=\textwidth,height=4.0cm]{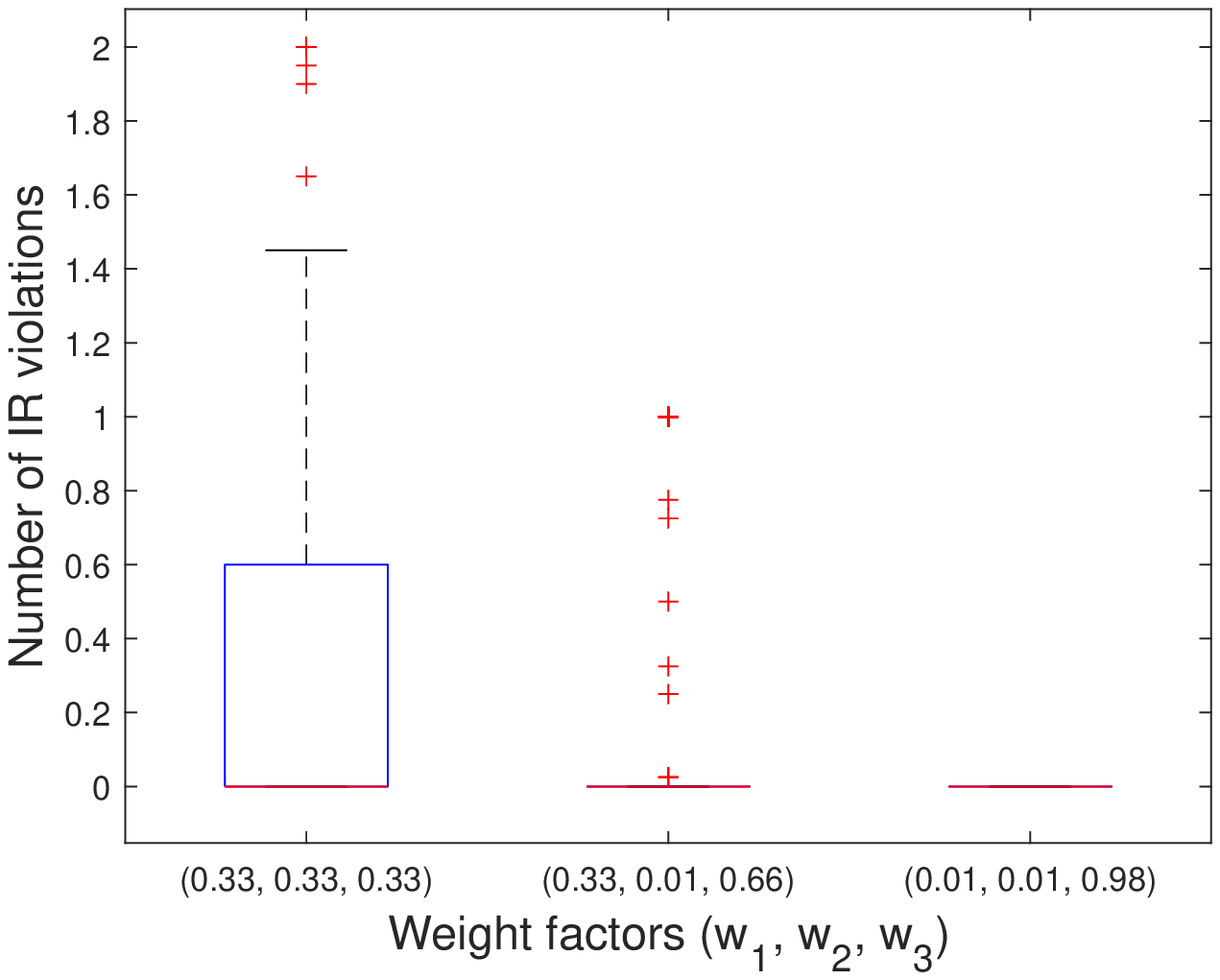}
         \caption{}
         \label{fig:weighting_factors_IR_Violations_Box}
     \end{subfigure}
     \hfill
     \begin{subfigure}[b]{0.3\textwidth}
         \centering
         \includegraphics[width=\textwidth,height=4.0cm]{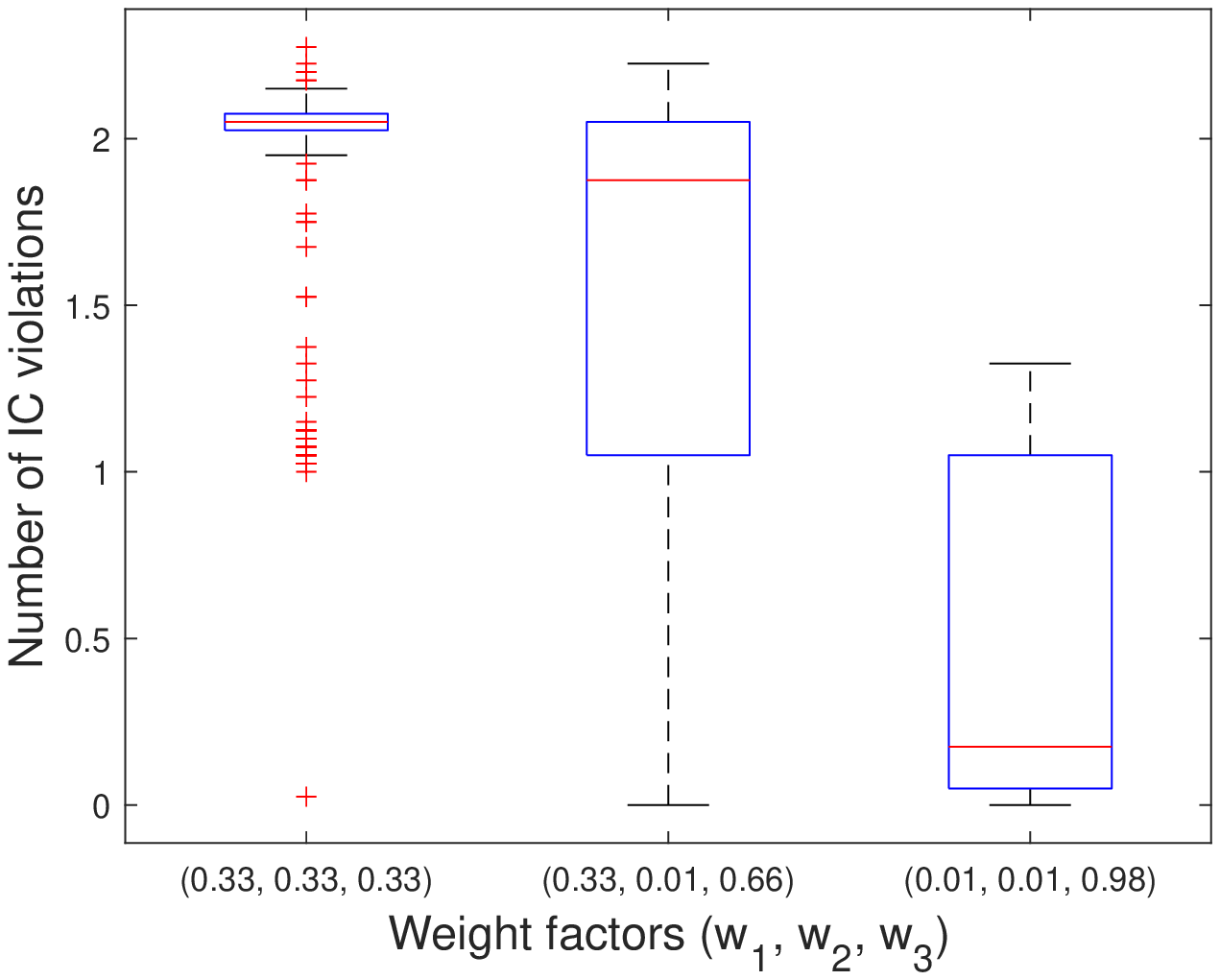}
         \caption{}
        \label{fig:weighting_factors_IC_Violations_Box}
     \end{subfigure}
     \hfill
     \begin{subfigure}[b]{0.3\textwidth}
         \centering
         \includegraphics[width=\textwidth,height=4.0cm]{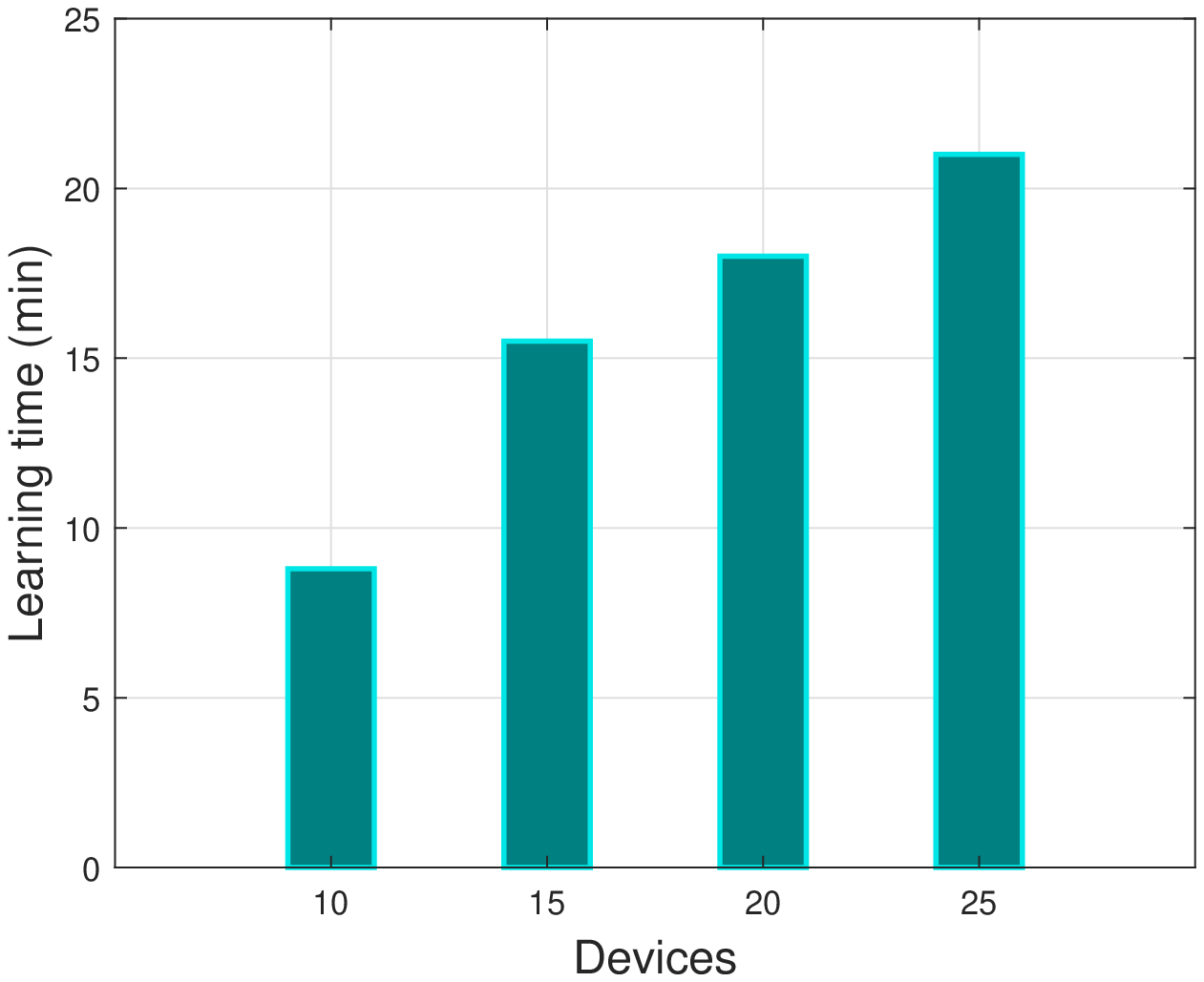}
         \caption{}
        \label{fig:Learning_time}
     \end{subfigure}
        \caption{(a) and (b) average number of IR and IC violations, respectively, when changing the weight factors. (c) Learning time.}
        \label{fig:weighting_factors_box}
\end{figure}

We also plot in Fig.~\ref{fig:weighting_factors_IR_Violations_Box} and Fig.~\ref{fig:weighting_factors_IC_Violations_Box} the average number of IR and IC violations when changing the weighting factors. Specifically, we measure the probability of IR and IC violations to be under 10 \% from episode 350 to episode 700. The results are shown in a box plot where on each box, the central mark indicates the median value and the bottom and top values of the box indicate the 25th and 75th percentiles, respectively. Fig.~\ref{fig:weighting_factors_IR_Violations_Box} validates the previous results that the IR violation rate diminishes as more weight is given to the IC term. Furthermore, we observe from Fig.~\ref{fig:weighting_factors_IC_Violations_Box} that when more weight is given to the IC term in the reward function, the majority of the IC violation rates are bellow 1 violation only on average. This results indicates that the derived solution to the contract problem is unlikely to violate the feasibility conditions.
We also plot in Fig.~\ref{fig:Learning_time} the time complexity of our DRL-based solution.
We observe that the MA-PDDQL has a linear complexity of $O(N)$ with respect to the number of devices. This is because at each iteration, the algorithm executes the PDDQL of each agent sequentially.

\subsubsection{Impact of the number of participants and the number of contract items}

\begin{figure*}[!h]
     \centering
     \begin{subfigure}[b]{0.3\textwidth}
         \centering
         \includegraphics[width=\textwidth,height=4.0cm]{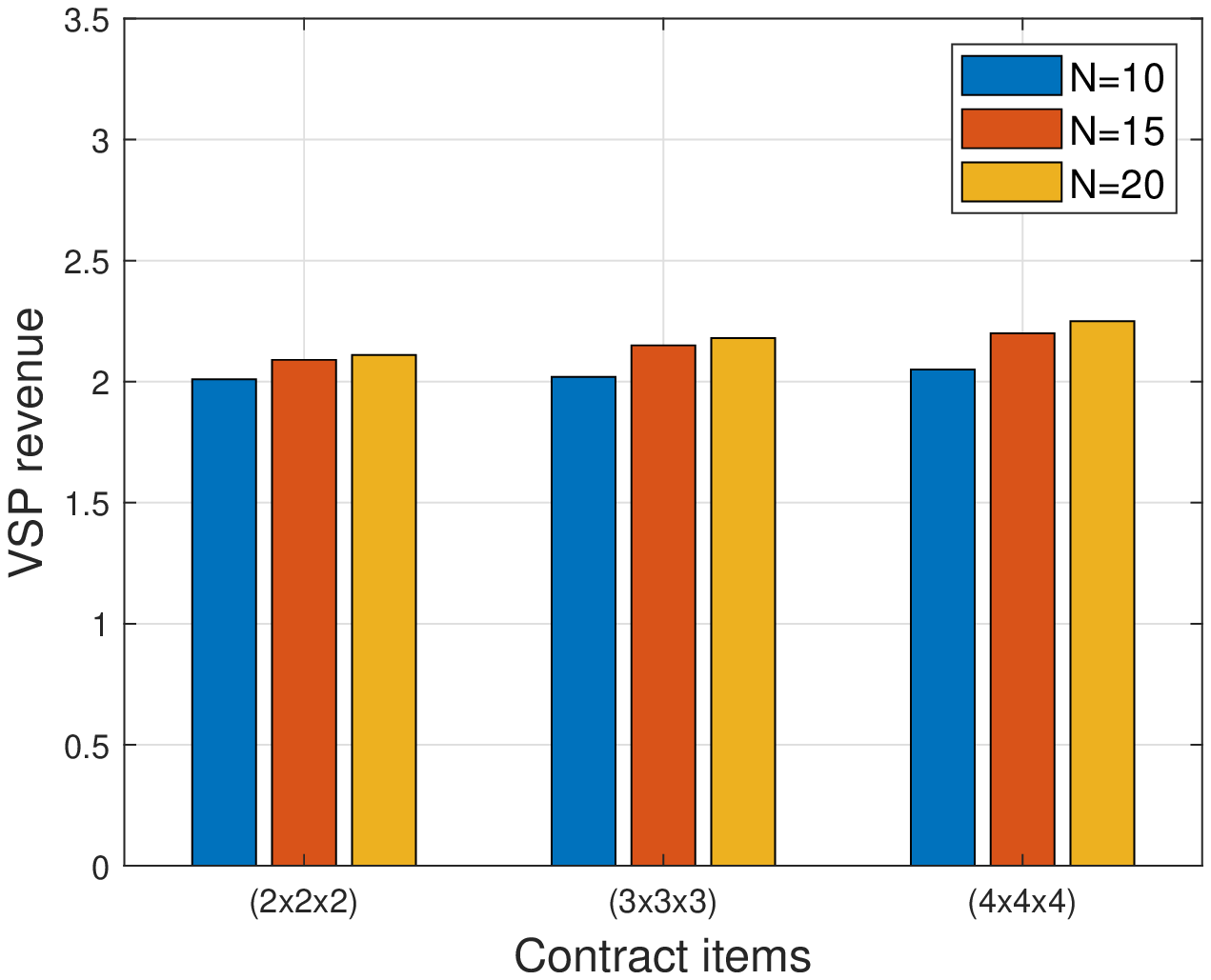}
         \caption{}
         \label{fig:nb_types_VSP_revenue}
     \end{subfigure}
     \hfill
     \begin{subfigure}[b]{0.3\textwidth}
         \centering
         \includegraphics[width=\textwidth,height=4.0cm]{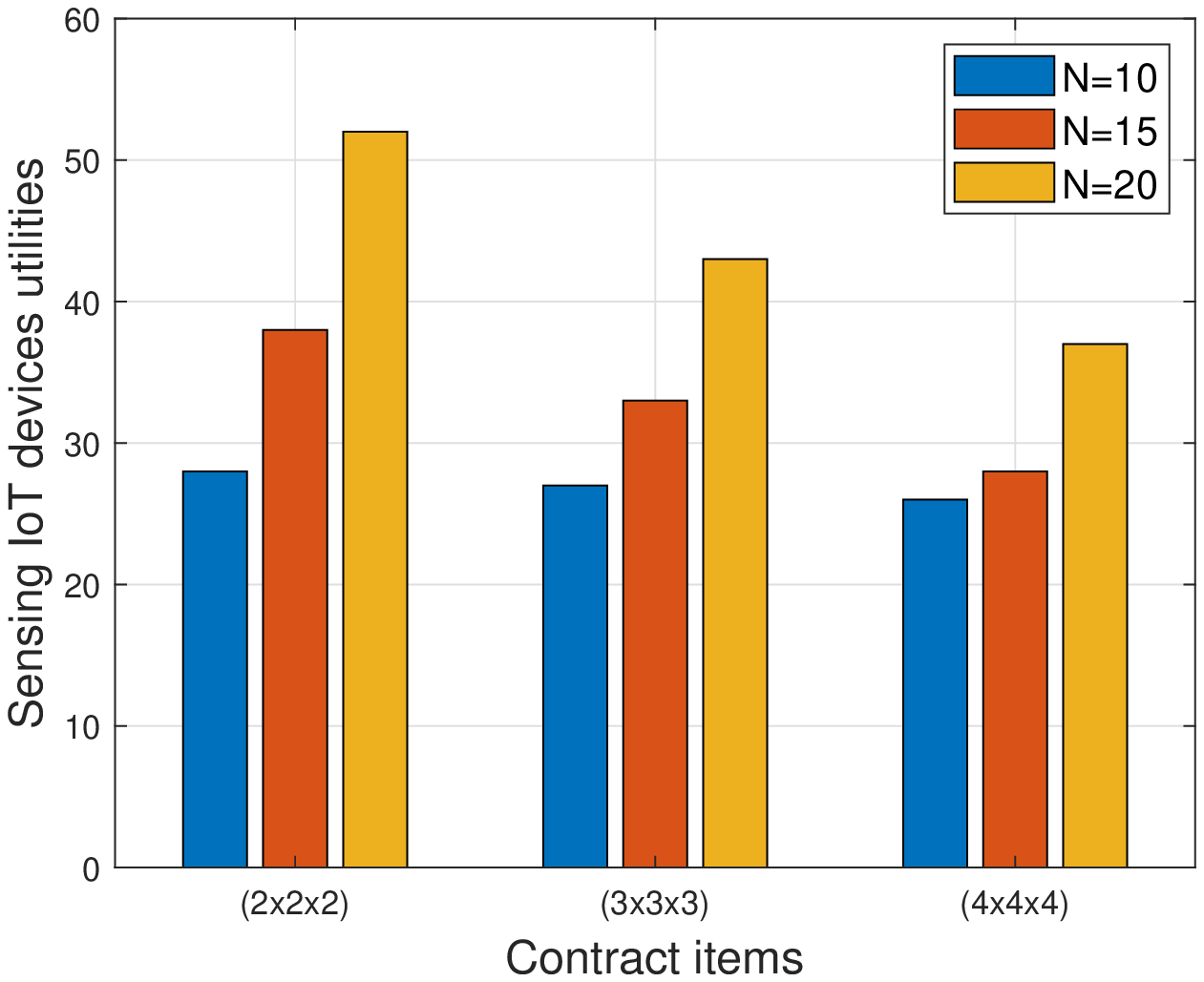}
         \caption{}
        \label{fig:nb_types_sensing_devices_utilities}
     \end{subfigure}
     \hfill
     \begin{subfigure}[b]{0.3\textwidth}
         \centering
         \includegraphics[width=\textwidth,height=4.0cm]{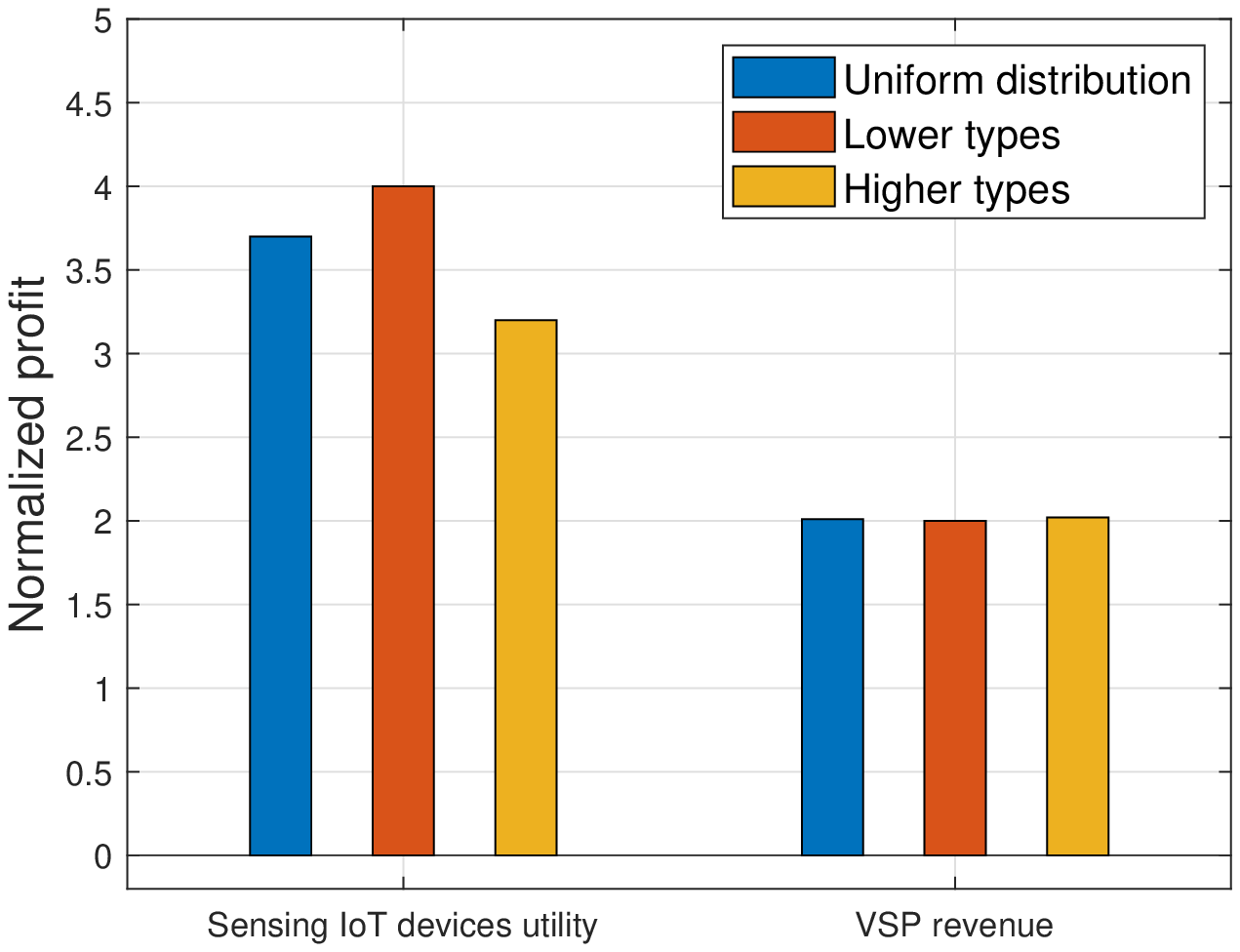}
    \caption{}
        \label{fig:distribution_change}
     \end{subfigure}
        \caption{(a) and (b)  VSP revenue and Utilities of the sensing IoT devices for different contract items. (c) Impact of changes in the distribution of the joint types.}
        \label{fig:nb_types_nb_devices}
\end{figure*}

In this experiment, we vary the number of combinations of the three-dimensional contract types and observe how the VSP revenue and the sensing IoT devices utilities change. The experiment is conducted on different number of participants, i.e., different $N$ sensing IoT devices, as shown in Fig.~\ref{fig:nb_types_VSP_revenue} and Fig.~\ref{fig:nb_types_sensing_devices_utilities}. We set three different scenarios for the number of contract items: 8(2x2x2), 27(3x3x3) and 64(4x4x4).
We observe that as the number of participating sensing IoT devices increases, the revenue of the VSP and the utility of the sensing IoT devices increase. This result is expected because more participating sensing IoT devices bring more semantic information to the VSP and hence the VSP gets higher utility.
Interestingly, we observe from Fig.~\ref{fig:nb_types_sensing_devices_utilities} that as the number of contract items increases, the utilities of the participating sensing IoT devices decreases. This is due to the fact that as the contract designer is able to derive more specific contract items for each participant based on their private information at a low level of precision, e.g., semantic value or AoI, it can extract more profit, which is reflected by the decrease in the utility of the participants. However, the profit that the VSP receives is marginal as observed from Fig.~\ref{fig:nb_types_VSP_revenue}, which is due the IR and IC constraints that have to be minimized. The algorithm adjust the contract bundles to satisfy IR and IC with less importance to the maximization of the VSP's revenue.

\subsubsection{Sensitivity to the distribution of types}

To further push our proposed framework to the limits, we train the model on a specific distribution of the types and then plug in other distribution and observe how the system reacts. Change of the distribution between training time and testing time is a common problem in DRL, see for example~\cite{Fang_ICIO_2022}, and its important to evaluate our framework for this change.
During training time, the joint type of each sensing IoT device is chosen uniformly from the different types' sets. However, at test time, we change the distribution of the joint type by changing the probabilities of each element in each set of types, i.e., $\Psi, \Lambda$ and $\Gamma$. The number of sensing IoT devices is set to 10. 
The following results are that of the augmented MA-PDDQL algorithm after convergence. The algorithm is given a set of tuples (semantic information sizes and prices) drawn from random values and the objective is to adjust the tuples to find an optimal solution that maximizes the revenue of the VSP and minimizes the number of IR and IC violations.
We consider three different scenarios. The first one is to consider the type of contract items drawn from the same distribution of the training time, i.e., uniform distribution. The second scenario is to consider the testing distribution drawn from a set with more weight on the lower types of the sets $\Psi, \Lambda$ and $\Gamma$. In the third scenario, we consider larger weights are given to higher types values.
Fig.~\ref{fig:distribution_change} shows the results of this experiment. 

We observe that the revenue of the VSP is marginally affected by the change of the distribution. However, the utilities of the sensing IoT devices when the lower types are giving larger weights are greater than that of the case of equal weights. Moreover, the utilities of the sensing IoT devices when the higher types are giving larger weight is less than that of the case of equal weights. We explain this behavior by observing that the DRL-based model is trained on types drawn from a uniform distribution. When faced with devices with lower types only (on all of the three dimensions), the model is not able to optimally minimize the gap between the cost and the price which makes the utilities of lower types devices higher. Similarly, this makes the utility of higher type devices less than that if the model was trained on the same distribution. However, note that the main objective of the VSP is not to minimize the utility of the sensing IoT devices. Instead its main objective, as shown in the objective function of $\mathcal{P}_1$ is to maximize its revenue while guaranteeing IR and IC, which is successfully achieved in all scenarios.
Moreover, our experiments show that the IR and IC violations remain low in all the cases. 
These results show the power of generalization of our model which can reach an optimal solution when facing newly observed scenarios.

\section{Conclusion}\label{section_conclusion}
In this paper, we design a semantic aware truthful mechanism for the Metaverse based on contract theory and MARL. Specifically, we design a two-layer Metaverse ecosystem where in the first layer, the VSP hires sensing IoT devices to obtain semantic information about the physical environment and render the digital twin. In the second layer, the VSP delivers the constructed digital twin to the Metaverse users. We then use contract theory to design the pricing bundles on both layers. We design a novel architecture for the contract in which the VSP interacts with the participants, i.e., sensing IoT devices or Metaverse users, to derive the optimal set of bundles. This interaction is conducted by using a new variant of MARL that we develop where the VSP creates DRL instances for each participant and sets the objective to maximize its revenue while minimizing the IR and IC violation rates. The simulation results show that our designed framework achieves good performance in terms of maximizing the profit of the VSP while not requiring several assumptions about the system model. In addition, when faced with a set of participants with types from a distribution different from the one they were trained on, our learning-based iterative contract is able to derive an optimal set of pricing bundles with minimal loss showing the generality of our framework to unobserved scenarios.
As a future work, it is interesting to explore the use of our proposed framework to solve bilinear optimization problems as they share a lot of similarities.
In addition, it is interesting to explore the strategy of joint optimization of the upstream layer and the downstream layer in a single MDP as it is expected to better help the VSP increase its profit and the profit of the Metaverse users.

\bibliographystyle{IEEEtran}
\bibliography{reference}

\end{document}